\documentclass[10pt,leqno]{amsart}
\usepackage{graphicx}
\usepackage{indentfirst,csquotes}

\usepackage{amssymb,amsthm,amsmath}
\usepackage{xcolor,paralist,hyperref,fancyhdr,etoolbox}

\usepackage{multirow}
\usepackage{multicol}
\usepackage{caption}
\usepackage{threeparttable}
\usepackage{algorithm}
\usepackage{algpseudocode}
\usepackage[T1]{fontenc}
\usepackage{lmodern}

\def \bb{\textcolor{blue}}

\usepackage{amsmath, amssymb, algorithm, algorithmicx, algpseudocode}

\newcommand{\beqs}{\vspace{0mm}\begin{eqnarray}}
\newcommand{\eeqs}{\vspace{0mm}\end{eqnarray}}
\newcommand{\barr}{\begin{array}}
\newcommand{\earr}{\end{array}}

\newcommand{\av}[0]{{\boldsymbol{a}}}
\newcommand{\bv}[0]{{\boldsymbol{b}}}

\newcommand{\fv}[0]{{\boldsymbol{f}} }

\newcommand{\sv}[0]{{\boldsymbol{s}}}

\newcommand{\xv}{\boldsymbol{x}}
\newcommand{\Xv}{\boldsymbol{X}}

\newcommand{\Tv}{\boldsymbol{T}}
\newcommand{\Wv}{\boldsymbol{W}}
\newcommand{\yv}{\boldsymbol{y}}

\newcommand{\Cv}{\boldsymbol{C}}

\newcommand{\thetav}{\boldsymbol{\theta}}

\newcommand{\swap}

\usepackage[numbers]{natbib}
\usepackage{booktabs}

\hypersetup{ colorlinks=true, linkcolor=black, filecolor=black, urlcolor=black }

\usepackage{lipsum}
\usepackage{caption}

\title{FastRef: Fast Prototype Refinement for Few-Shot Industrial Anomaly Detection}

\author{
Long Tian\textsuperscript{1,*},
Yufei Li\textsuperscript{1},
Yuyang Dai\textsuperscript{1},
Wenchao Chen\textsuperscript{2},
Xiyang Liu\textsuperscript{1},
Bo Chen\textsuperscript{2},
}

\thanks{\textsuperscript{1} School of Computer Science and Technology, Xidian University, Xi'an, China}
\thanks{\textsuperscript{1} School of Electronic Engineering, Xidian University, Xi'an, China}
\thanks{*Corresponding author: Long Tian (tianlong@xidian.edu.cn)}

\date{\today}
\begin{document}


\begin{abstract}
Few-shot industrial anomaly detection (FS-IAD) presents a critical challenge for practical automated inspection systems operating in data-scarce environments. While existing approaches predominantly focus on deriving prototypes from limited normal samples, they typically neglect to systematically incorporate query image statistics to enhance prototype representativeness. To address this issue, we propose FastRef, a novel and efficient prototype refinement framework for FS-IAD.
Our method operates through an iterative two-stage process: (1) characteristic transfer from query features to prototypes via an optimizable transformation matrix, and (2) anomaly suppression through prototype alignment. The characteristic transfer is achieved through linear reconstruction of query features from prototypes, while the anomaly suppression addresses a key observation in FS-IAD that unlike conventional IAD with abundant normal prototypes, the limited-sample setting makes anomaly reconstruction more probable. Therefore, we employ optimal transport (OT) for non-Gaussian sampled features to measure and minimize the gap between prototypes and their refined counterparts for anomaly suppression. For comprehensive evaluation, we integrate FastRef with three competitive prototype-based FS-IAD methods: PatchCore, FastRecon, WinCLIP, and AnomalyDINO. Extensive experiments across four benchmark datasets of MVTec, ViSA, MPDD and RealIAD demonstrate both the effectiveness and computational efficiency of our approach under 1/2/4-shots.
\end{abstract} 
\maketitle

\section{Introduction}
\label{sec:introduction}

Industrial Anomaly Detection (IAD) aims to automatically identify defects on product surfaces \cite{liu2024deep} and has been attracting tremendous attention \cite{zhao2023omnial,you2022unified,lu2023hierarchical}. The fragmented nature of industrial anomalies ranging from subtle bruises to obvious breakages, with varying appearances and scales \cite{roth2022towards} , making it difficult for fully supervised methods \cite{he2017mask,kamat2020anomaly}. Therefore, unsupervised IAD methods, trained with massive normal images, have been developed \cite{liu2023simplenet,lu2024hierarchical}. In practice, it is not always possible to obtain a large number of normal images for different products, making existing methods less 
in a low data regime at test time \cite{huang2022registration}.

\begin{figure}[t]
\centering
\centerline{\includegraphics[width=9cm]{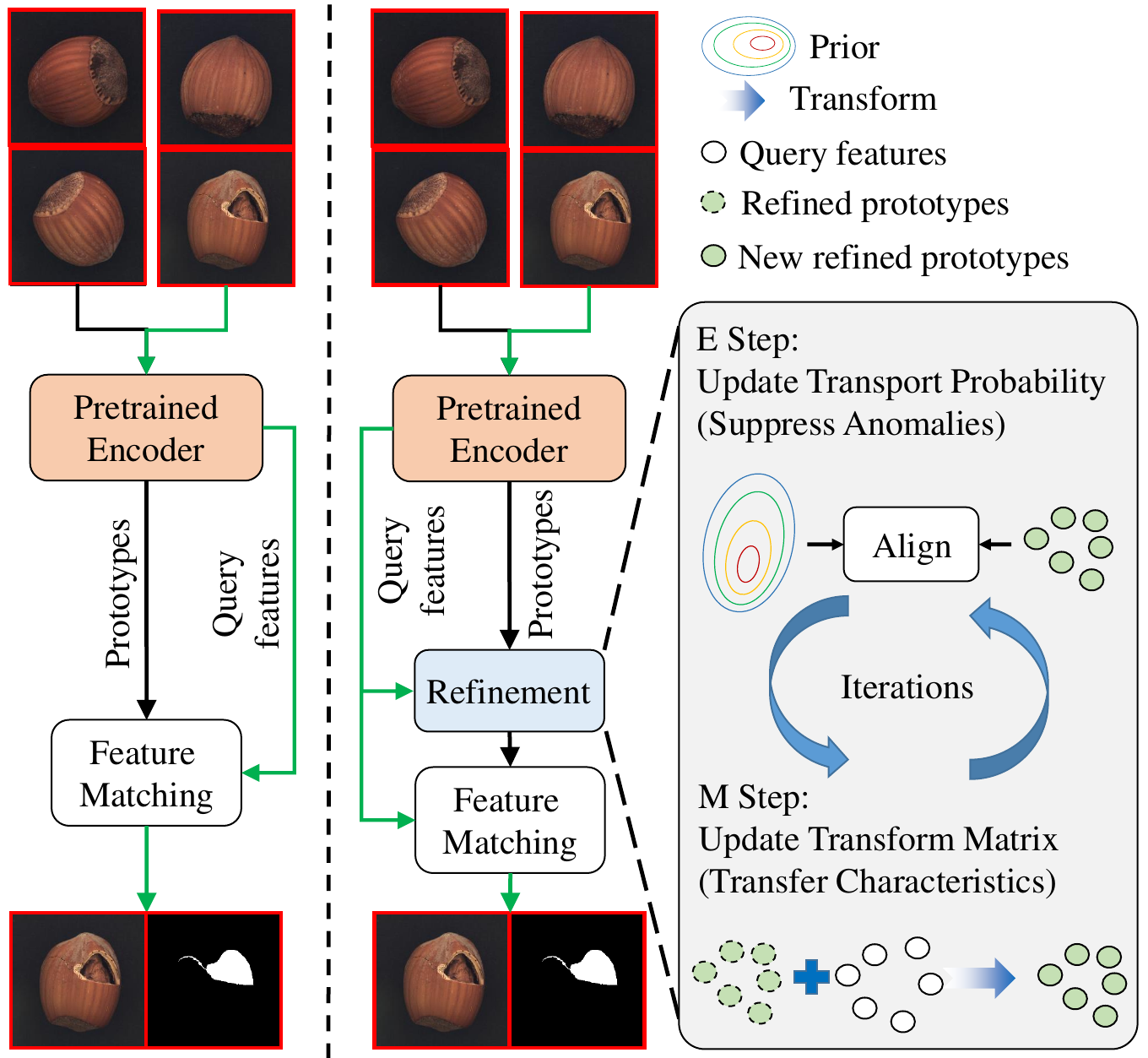}}
\caption{\textbf{Left:} The pipeline of traditional prototype-oriented few-shot IAD methods. \textbf{Right:} Our proposed framework of FastRef, it refines prototypes with the current query features by iterating between suppressing anomalies and transferring characteristics. The improved anomaly detection performance could be obtained thanks to the well refined prototypes.
}
\label{fig:motivation}
\end{figure}

To tackle this challenge, few-shot learning \cite{snell2017prototypical,sung2018learning,wang2020generalizing} has been introduced to unsupervised IAD, allowing the development of a common model
generalizing to new products with only a few normal training images, such as 1-shot per product. This new paradiagm is known as few-shot (unsupervised) IAD \cite{huang2022registration}, 
and mainly involves prototype-oriented methods \cite{fang2023fastrecon,jeong2023winclip,santos2023optimizing}. At training time, these methods typically use the statistics of a few normal training (support) images to construct normal prototypes. At test time, anomaly scores are computed by measuring the differences between test (query) images and normal prototypes using some well-defined measurement functions. Anomalies are detected by comparing the anomaly score against predefined thresholds. Especially, \cite{fang2023fastrecon} further uses query features to quickly refine prototypes at test time. However, we find that point-to-point regularization as in \cite{fang2023fastrecon} does significantly limits the ability to transfer characteristics from query images to prototypes. Additionally, meta learning based few-shot IAD (FS-IAD) methods \cite{wu2021learning,huang2022registration} have been introduced to achieve fast generalization, whose performance is verified to be far behind of prototype-oriented methods \cite{xie2023pushing}.

Observing the fact that the characteristics of the query images have not been fully explored at test time previously \cite{fang2023fastrecon,jeong2023winclip,santos2023optimizing} either from the data perspective \cite{jeong2023winclip,santos2023optimizing} or from the optimization perspective \cite{fang2023fastrecon}, which may result in suboptimal prototype refinement
, particularly when the available normal training images are extremely limited. Therefore, we propose a novel and fast prototype refinement framework called FastRef, as shown in Fig. \ref{fig:motivation}, to enhance the representativeness of prototypes by transferring characteristics from query images while suppressing anomalies present in query images, making the prototypes more robust and generalizable. 
At test time, we formulate prototype refinement with query images as a nested optimization process balancing anomaly suppression and characteristic transfer, with a transport probability and transform matrix capturing these behaviors, respectively. We then introduce an efficient iterative-based algorithm to solve the transport probability and transform matrix. As for anomaly suppression, we use Sinkhorn algorithm \cite{cuturi2013sinkhorn}, an entropy-regularized optimal transport, to align the distribution between the prototypes and their refined counterparts, ensuring that the refined prototypes are unaffected by anomalies. When it comes to characteristic transfer, we employ gradient descent to maximize the transferability from query images to the refined prototypes in a closed form. Finally, the proposed framework of FastRef is integrated into three popular and recently proposed prototype-oriented few-shot IAD methods, PatchCore \cite{roth2022towards}, FastRecon \cite{fang2023fastrecon}, WinCLIP \cite{jeong2023winclip}, and AnomalyDINO \cite{damm2025anomalydino}, to enhance the representativeness of their prototypes. Our model consistently improves the performance of the three existing methods by significant margins across four widely used datasets, including MVTec \cite{bergmann2019mvtec}, ViSA \cite{zou2022spot}, MPDD \cite{jezek2021deep}, and RealIAD \cite{wang2024real}. Additionally, we find that proper initialization of transport probability achieves optimal efficiency-performance trade-offs and only requires less than 10 inner loops by Sinkhorn algorithm in practice.
The main contributions can be summarized as follows:

\begin{itemize}
    \item We present FastRef, a fast prototype refinement framework, to transfer characteristics of query features while suppressing anomalies in query features, which can be easily integrated with other prototype-oriented methods.
    \item We formulate FastRef as a nested optimization problem and introduce an efficient optimization algorithm to solve the problem precisely in a closed form during inference.
    \item The experimental results confirm that FastRef is effective in improving FS-IAD performance while maintaining real-time efficiency on the datasets of MVTec, ViSA, MPDD, and RealIAD.
\end{itemize}

\section{Related Works}
\label{sec:related}

\subsection{Anomaly Detection}
Industrial anomaly detection (IAD) involves handling training images that exclusively consist of normal data,
primarily falling into two categories of reconstruction-based methods \cite{he2023diad,wyatt2022anoddpm,gong2019memorizing,you2022unified,lu2023hierarchical} and memory-based methods \cite{roth2022towards,cohen2020sub,defard2021padim}. Reconstruction-based methods are trained exclusively with normal images on the premise that anomalies will yield significantly higher reconstruction errors \cite{gong2019memorizing}. To address shortcut, Transformer-based architectures \cite{you2022unified,lu2023hierarchical} and diffusion-based training strategies \cite{wyatt2022anoddpm,roth2022towards} have been developed concurrently. Memory-based methods take full advantages of pre-trained features to improve detection performance. However, both approaches tend to overfit when the number of normal training images per product is limited \cite{huang2022registration}.

\subsection{Few-shot Anomaly Detection}
Few-shot IAD has developed to address the demand for rapid manufacturing changeovers, with research mainly divided into prototype-oriented methods \cite{santos2023optimizing,xie2023pushing,fang2023fastrecon,jeong2023winclip,gu2024anomalygpt} and meta-learning based methods \cite{wu2021learning,huang2022registration}. Prototype-oriented methods usually use pre-trained features to construct normal prototypes from only a few normal training images, with a focus on obtaining generalizable prototypes.
\cite{xie2023pushing} develop graph Swin-Transformer \cite{liu2021swin} to extract isometric-invariant visual features. \cite{jeong2023winclip,li2024promptad} turn to use Large Language Models (LLMs) \cite{radford2021learning} to create powerful prototypes.
\cite{fang2023fastrecon} leverage query image characteristics to enhance prototype representativeness. Although these methods efficiently build generalizable prototypes, they still lack a systematic way for refining prototypes properly at test time.

\section{Background}
\label{sec: background}

\subsection{Task Formulation}
We formally define the one-class IAD task in a low data regime, adhering to the standard few-shot learning. The model is fine-tuned using $\rm{k}$ normal support images $\xv_{1:{\rm{k}}}^{\rm{s}}$ and predicts whether the ${\rm{t}}$-th query image $\xv_{\rm{t}}^{\rm{q}}$ is anomalous at both the pixel and image levels.
In this paper, we fully leverage the pre-trained backbone $f_{\thetav^*}$, parameterized by $\thetav^*$, to extract features from both support and query images, computed as follows:
\begin{align}
\begin{split}
    \fv_{1:{\rm{k}}\times{\rm{h}}\times{\rm{w}}}^{\rm{s}}={\rm{flatten}}[f_{\thetav^*}(\xv_{1:{\rm{k}}}^{\rm{s}})], \quad \fv_{\rm{t}}^{\rm{q}}=f_{\thetav^*}(\xv_{\rm{t}}^{\rm{q}})
\end{split}
\end{align}
where ${\rm{flatten}}[\cdot]$ is an operation that converts a 2-D feature map into a 1-D vector, 
and 
$\fv_{\rm{t}}^{\rm{q}} \in \mathbb{R}^{{{\rm{h}}}\times{{\rm{w}}}\times{\rm{c}}}$. In practice, normal features tend to be redundant, we use compression techniques like Coreset \cite{sener2017active} to construct prototypes $\boldsymbol{\mathcal{M}}_{\rm{s}} \in \mathbb{R}^{\alpha \times{{\rm{k}}}\times{{\rm{h}}}\times{{\rm{w}}} \times {\rm{c}}}$ by selecting the most representative normal features from $\fv_{1:{{\rm{k}}}\times{{\rm{h}}}\times{{\rm{w}}}}^{\rm{s}}$ with a downsampling ratio $\alpha \in (0,1)$. For simplicity, we denote ${{\rm{m}}}={{\rm{h}}}\times{{\rm{w}}}$ and ${{\rm{n}}}={\alpha}\times{{\rm{k}}}\times{{\rm{h}}}\times{{\rm{w}}}$ for the reminder of this paper.

\subsection{Optimal Transport}
\label{sec: ot}
Although optimal transport (OT) has a rich theory, we limit our discussion to OT for discrete distributions and refer the readers to \cite{peyre2019computational} for more details. Let us consider $p$ and $q$ as two discrete probability distributions on the arbitrary space $\Xv \subset \mathbb{R}^{\rm{c}}$, which can be formulated as $p=\sum_{{\rm{i}}=1}^{\rm{m}}a_{\rm{i}} \delta_{\xv_{\rm{i}}}$, and $q=\sum_{{\rm{j}}=1}^{\rm{n}}b_{\rm{j}}\delta_{\yv_{\rm{j}}}$. In this case, $\av \in \Sigma^{\rm{m}}$ and $\bv \in \Sigma^{\rm{n}}$, where $\Sigma^{\rm{m}}$ denotes the probability simplex of $\mathbb{R}^{\rm{m}}$. The OT distance between $\av$ and $\bv$ is defined as:
\begin{align}
\begin{split}
    {\rm{OT}}(p,q)={\rm{min}}_{\Tv \in U(p,q)} \left \langle \Tv, \Cv \right \rangle
\end{split}
\end{align}
where $\left \langle \cdot, \cdot \right \rangle$ denotes the Frobenius dot-product, $\Cv \in \mathbb{R}^{{\rm{m}}\times{\rm{n}}}_{\geq 0}$ is the cost matrix with element $C_{{\rm{i}},\rm{j}}=\Cv(\xv_{\rm{i}},\yv_{\rm{j}})$, and $\Cv(\cdot,\cdot)$ is usually distance function like Euclidean or cosine. $\Tv \in \mathbb{R}^{{\rm{m}}\times{\rm{n}}}_{\textgreater 0}$ is the doubly stochastic transport probability such that $U(p,q):=\{\Tv|\sum_{{\rm{i}}=1}^{\rm{m}}\Tv_{\rm{i},\rm{j}}=\bv_{\rm{j}}, \sum_{{\rm{j}}=1}^{\rm{n}} \Tv_{\rm{i},\rm{j}}=\av_{\rm{i}}\}$. To relax the time-consuming problem when optimizing the OT distance, \cite{cuturi2013sinkhorn} introduced the entropic regularization, $H=-\sum_{\rm{i},\rm{j}} \Tv_{\rm{i},\rm{j}} {\rm{ln}} \Tv_{\rm{i},\rm{j}}$, leading to the widely-used Sinkhorn algorithm for discrete OT problems.

\section{Method}
\label{sec: method}

Our FastRef is presented as follows:
We first formulate our fast prototype refinement framework as a nested optimization problem between characteristic transfer and anomaly suppression in Sec. \ref{sec: formulation}. Then we introduce an OT-based efficient iterative updating rule for transformation matrix and transport probability in Sec. \ref{sec: solution}. Finally, we implement the improved anomaly detection through reconstruction using query features and the refined prototypes in Sec. \ref{sec: detection}.

\subsection{A Nested Optimization for Modeling FastRef}
\label{sec: formulation}

We aim to refine prototypes by transferring characteristics while suppressing anomalies from query images. We propose a nested optimization for modeling such behaviors as:
\begin{align} \small \label{eq: obj}
\begin{split}
    \Wv^*, \Tv^* = \mathop{\arg\min}\limits_{\Wv, \Tv} {\rm{dis}}(\fv_{{\rm{t}}}^{{\rm{q}}},\Wv \boldsymbol{\mathcal{M}}_{\rm{s}})+\lambda {\rm{OT}}(p, q)
\end{split}
\end{align}
where ${\rm{dis}}(\cdot,\cdot)$ is the point-wise distance between two sets such as Euclidean, with its form dependent on the follow-up applications in Sec. \ref{sec: application}. $\Wv \in \mathbb{R}^{{\rm{m}} \times {\rm{n}}}$ is the transform matrix used to transfer the characteristics of query features $\fv_{\rm{t}}^{\rm{q}}$ 
for obtaining refined prototypes $\boldsymbol{{\mathcal{M}}}^{\rm{w}}_{\rm{s}}=\Wv \boldsymbol{\mathcal{M}}_{\rm{s}}$ 
, where $\boldsymbol{{\mathcal{M}}}^{\rm{w}}_{\rm{s}} \in \mathbb{R}^{{\rm{m}}\times{\rm{c}}}$. Notably, different from prototype refinement that takes place along the feature dimension, we refer to our transform as composition refinement, the meaning of which is reconstructing query features by selecting items from the normal prototypes. 
$\Tv \in \mathbb{R}^{{\rm{n}}\times{\rm{m}}}_{\textgreater 0}$ is the transport probability discussed in OT background in Sec. \ref{sec: background}. It is used to suppress anomalies potentially present in query features $\fv_{\rm{t}}^{\rm{q}}$ by minimizing ${\rm{OT}}(p,q)$. $p$ and $q$ are distributions expanded by $\boldsymbol{\mathcal{M}}^{\rm{w}}_{\rm{s}}$ and $\boldsymbol{{\mathcal{M}}}_{\rm{s}}$, where $p=\sum_{{\rm{i}}=1}^{\rm{m}} \frac{1}{\rm{m}} \delta_{\boldsymbol{\mathcal{M}}^{\rm{w}}_{{\rm{s}},{\rm{i}}}}$, $q=\sum_{{\rm{j}}=1}^{\rm{n}} \frac{1}{\rm{n}} \delta_{\boldsymbol{{\mathcal{M}}}_{{\rm{s}},{\rm{j}}}}$, and $\boldsymbol{\mathcal{M}}_{{\rm{s}},{\rm{i}}},\boldsymbol{{\mathcal{M}}}^{\rm{w}}_{{\rm{s}},{\rm{j}}} \in \mathbb{R}^{\rm{c}}$.
Minimizing the second term in Eq. \ref{eq: obj} makes the refined prototypes $\boldsymbol{{\mathcal{M}}}^{\rm{w}}_{\rm{s}}$ much more similar with the original prototypes since $\boldsymbol{\mathcal{M}}_{\rm{s}}$ is normal, thus suppressing potential anomalies in query features. With the constraint of $\sum_{{\rm{j}}} \Tv_{{\rm{i}},{\rm{j}}}=\frac{1}{{\rm{m}}}$, although the transport probability may pay attention to the anomaly features in $\boldsymbol{{\mathcal{M}}}^{\rm{w}}_{\rm{s}}$, in order to minimize the OT term in Eq. \ref{eq: obj}, in the next round of optimization on $\Wv$, theoretically, it may ignore those anomaly features by adjusting the corresponding weights in $\boldsymbol{W}$. Therefore, the anomalies could be suppressed. The details of updating between $\Wv$ and $\Tv$ can be found in Sec. \ref{sec: solution}. On the contrary, if we only adopt point-to-point distance between $\fv_{{\rm{t}}}^{{\rm{q}}}$ and $\Wv \boldsymbol{\mathcal{M}}_{\rm{s}}$ as shown in the first term of Eq. \ref{eq: obj}. Then the refined prototypes may be misled by anomalies in $\fv_{{\rm{t}}}^{{\rm{q}}}$ so that the distance between $\fv_{{\rm{t}}}^{{\rm{q}}}$ and $\Wv \boldsymbol{\mathcal{M}}_{\rm{s}}$ could be smaller. Namely, the anomalies can not be removed without the OT objective.
Additionally, 
our OT-based regularization does not rely on Gaussian assumption,
making it more adaptable for real-world few-shot IAD. $\lambda$ is the balanced coefficient.
We then use Sinkhorn algorithm \cite{cuturi2013sinkhorn} and 
re-write
the second term in Eq. \ref{eq: obj} as follows:
\begin{align} \label{eq: ot}
\begin{split}
    {\rm{OT}}_{\epsilon}(p_{\rm{s}},q_{\rm{s}}):=\sum_{{\rm{i}},{\rm{j}}}^{{\rm{m}},{\rm{n}}} \Cv_{{\rm{i}},{\rm{j}}} \Tv_{{\rm{i}},{\rm{j}}} - \epsilon \sum_{{\rm{i}},{\rm{j}}}^{{\rm{m}},{\rm{n}}} -\Tv_{{\rm{i}},{\rm{j}}} {\rm{ln}} \Tv_{{\rm{i}},{\rm{j}}}
\end{split}
\end{align}
where $\epsilon \textgreater 0$, $\Cv \in \mathbb{R}^{{m}\times{n}}_{\geq 0}$ 
is typically formulated with some simple distance functions ${\rm{dis}}(\cdot,\cdot)$ like Euclidean and cosine, as detailed in Sec. \ref{sec: application}.
Recalling that $\Tv \in \mathbb{R}^{{\rm{m}}\times{\rm{n}}}_{\textgreater 0}$ must satisfy $U(p,q):=\{\sum_{{\rm{j}}=1}^{\rm{n}} \Tv_{{\rm{i}},{\rm{j}}}=\frac{1}{\rm{m}}, \sum_{{\rm{i}}=1}^{\rm{m}} \Tv_{{\rm{i}},{\rm{j}}}=\frac{1}{\rm{n}}\}$, where $\Tv_{{\rm{i}},{\rm{j}}}$ denotes the transport probability between the ${\rm{i}}$-th refined prototype and the ${\rm{j}}$-th prototype, serving as an upper-bounded positive metric. Consequently, $\Tv$ naturally weights the importance of each refined prototype in relation the set of original (normal) prototypes.

\begin{figure}[t]
\centering
\centerline{\includegraphics[width=12cm]{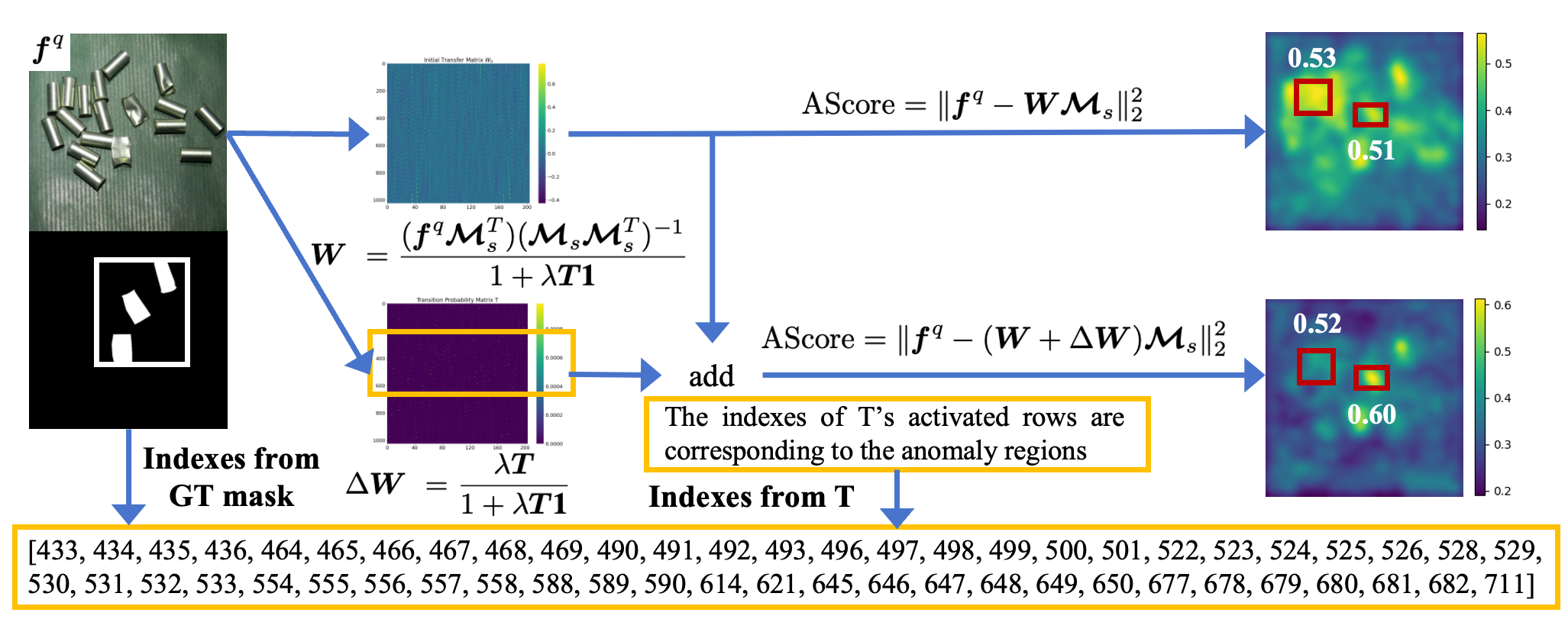}}
\caption{\textcolor{black}{The roles of transformation matrix $\boldsymbol{W}$ and transport probability $\boldsymbol{T}$. \textbf{Top:} $\boldsymbol{W}$-based $\fv_{{\rm{t}}}^{{\rm{q}}}$ reconstruction restores normal features but may propagate anomalies, causing similar anomaly score (AScore) in defective and normal regions. \textbf{Bottom:} $\boldsymbol{T}$-adjusted $\boldsymbol{W}$ enables effective anomaly suppression, making defective regions' AScore distinguishable. The indexes between GT mask and $\boldsymbol{T}$ show that $\boldsymbol{T}$ primarily activates on anomalous areas 
, which is crucial for enhancing prototype representativeness.}
}
\label{fig:rule}
\end{figure}

\subsection{An Efficient Algorithm for Solving FastRef}
\label{sec: solution}

Observing the nested optimization modeling in Eq. \ref{eq: obj}, it is obvious that the optimal parameters for the transport probability $\Tv^*$ and the transform matrix $\Wv^*$ are interdependent. 
This motivates us to develop an efficient algorithm for iterative solving.
At the $l$-th iteration, for the purpose of anomaly suppression, we keep the transform matrix $\Wv_{l}$ fixed and update the transport probability $\Tv_{l}$ by minimizing the second term of Eq. \ref{eq: obj} using Sinkhorn algorithm \cite{cuturi2013sinkhorn} to derive $\Tv_{l+1}$. For characteristic transfer, we keep the transport probability $\Tv_{l+1}$ fixed and update transform matrix $\Wv_{l}$ by minimizing Eq. \ref{eq: obj}, denoted as $\mathcal{L}(\fv_{\rm{t}}^{\rm{q}},\boldsymbol{\mathcal{M}}_{\rm{s}};\Wv,\Tv)$, using gradient descent as $\Wv_{l+1}=\Wv_{l}+\beta \frac{\partial \mathcal{L}(\fv_{\rm{t}}^{\rm{q}},\boldsymbol{\mathcal{M}}_{\rm{s}};\Wv_{l},\Tv_{l+1})}{\partial \Wv_{l}}$.
After ${\rm{L}}$ steps, the optimal refined prototypes can be expressed as $\boldsymbol{{\mathcal{M}}}^{\rm{w}*}_{\rm{s}}=\Wv^* \boldsymbol{\mathcal{M}}_{\rm{s}}$, where $\Wv^*=\Wv_{\rm{L}}$ and $\Tv^*=\Tv_{\rm{L}}$. 
Moreover, a closed-form solution for updating $\Wv_{l+1}$ can be arrived when we choose ${\rm{dis}}(\cdot,\cdot)$ in Eq. \ref{eq: obj} as Euclidean or cosine distance, and we have:
\begin{align} \label{eq: updatew}
\begin{split}
    \boldsymbol{W}_{l+1}=\frac{(\boldsymbol{f}_t^q\boldsymbol{\mathcal{M}}_s^T+\lambda \boldsymbol{T}_l\boldsymbol{\mathcal{M}}_s\boldsymbol{\mathcal{M}}_s^T)(\boldsymbol{\mathcal{M}}_s\boldsymbol{\mathcal{M}}_s^T)^{-1}}{1+\lambda \boldsymbol{T}_l \cdot \textbf{1}}
\end{split}
\end{align}
where $\boldsymbol{\mathcal{M}}_{\rm{s}} \boldsymbol{\mathcal{M}}_{\rm{s}}^T \in \mathbb{R}^{{\rm{n}}\times{\rm{n}}}$, we could compute the inverse of the matrix in advance and prepare to reuse it. In order to take a balance between accuracy and inference speed, we use Coreset technique \cite{sener2017active} to set ${\rm{n}}$ small. Additionally, updating $\boldsymbol{T}_{l}$ at each iteration is another important computation bottleneck, we find that proper initialization with $\boldsymbol{W}_0=(\fv_{\rm{t}}^{\rm{q}} \boldsymbol{\mathcal{M}}_{\rm{s}}^T)(\boldsymbol{\mathcal{M}}_{\rm{s}}^T \boldsymbol{\mathcal{M}}_{\rm{s}})^{-1}$ achieves optimal efficiency-performance trade-offs and only requires about $10$ inner loops by Sinkhorn algorithm in practice. We also find that ${\rm{L}}=2$ yields promising few-shot IAD results in practice. All of these indicate that FastRef optimization can satisfy real-time requirements. 
The detailed derivation of Eq. \ref{eq: updatew} is presented in the follow-up Appendix.

\begin{algorithm}[t]
\caption{Few-shot IAD of FastRef}
\label{alg: few-shot iad}
\begin{algorithmic}[1]
\Require Initial transform matrix $\Wv_0$, maximum iteration number ${\rm{L}}$, prototypes $\boldsymbol{\mathcal{M}}_{\rm{s}}$, the ${{\rm{t}}}$-th query features $\fv_{{\rm{t}}}^{{\rm{q}}}$
\State Calculate $q$ using $\boldsymbol{\mathcal{M}}_{\rm{s}}$
\State \textbf{For} {$l = 0$ to ${\rm{L}}-1$} \textbf{do}
\State \quad Calculate $p$ using $\Wv_l \boldsymbol{\mathcal{M}}_{\rm{s}}$
\State \quad Update $\Tv_{l+1}$ by minimizing $\text{OT}_{\rm{\epsilon}}(p, q)$ while fix $\Wv_l$
\State \quad Update $\Wv_{l+1}$ according to Eq. \ref{eq: updatew} while fix $\Tv_{l+1}$
\State Set $\Wv^* = \Wv_{\rm{L}}$, $\boldsymbol{\mathcal{M}}^{\rm{w}*}_{\rm{s}}=\Wv^*\boldsymbol{\mathcal{M}}_{\rm{s}}$
\State Implement few-shot IAD according to Eq. \ref{eq: det}.
\end{algorithmic}
\end{algorithm}

\subsection{Anomaly Detection through Reconstruction}
\label{sec: detection}

Once the optimal refined prototypes $\boldsymbol{{\mathcal{M}}}^{\rm{w}*}_{\rm{s}}$ are obtained according to Sec. \ref{sec: solution}, the anomaly score map $\sv$ for the ${\rm{t}}$-th query image $\xv_{\rm{t}}^{\rm{q}}$ can be defined by calculating the similarities between $\boldsymbol{{\mathcal{M}}}^{\rm{w}*}_{\rm{s}}$ and $\fv_{\rm{t}}^{\rm{q}}$, the features of $\xv_{\rm{t}}^{\rm{q}}$, as follows:
\begin{align} \label{eq: det}
\begin{split}
    \sv_{\rm{j}}:={\rm{min}}_{\boldsymbol{r} \in \boldsymbol{{\mathcal{M}}}^{\rm{w}*}_{\rm{s}}} {\rm{dis}}(\fv_{\rm{t},\rm{j}}^{\rm{q}}, \boldsymbol{r}), \quad {\rm{j}}=1,...,{\rm{m}}
\end{split}
\end{align}
where the distance function ${\rm{dis}}(\cdot,\cdot)$ is consistent with that in Eq. \ref{eq: obj}. For image-level anomaly detection, we represent the maximum score $s^*$ among all values in $\sv \in \mathbb{R}^{\rm{m}}$ as $s^*={\rm{max}}_{{\rm{j}} \in [1,{\rm{m}}]}\sv_{\rm{j}}$. For pixel-level localization, we first upscale the anomaly score map $\sv$ using bi-linear interpolation to match the original input resolution. We smooth the score map using a Gaussian kernel following \cite{roth2022towards}
. Additionally, the pseudo-code is provided in Alg. \ref{alg: few-shot iad}.

\subsection{Convergence Analysis}
The iterative-based algorithm for solving FastRef in Sec. \ref{sec: solution} is convergent and we have the following inequality:
\begin{align*}
\begin{split}
    \mathcal{L}(\boldsymbol{W}_{l+1},\boldsymbol{T}_{l+1}) & \leq \mathcal{L}(\boldsymbol{W}_l,\boldsymbol{T}_l)-\frac{\mu}{2} \|\boldsymbol{W}_{l+1}-\boldsymbol{W}_l\|_{\rm{F}}^2 
\end{split}
\end{align*}
where $\mathcal{L}(\boldsymbol{W},\boldsymbol{T})$ represents the objective in Eq. \ref{eq: obj}. The detailed derivation can be found in the followup Appendix.

\section{Application to Prototype-oriented Methods}
\label{sec: application}

we apply the proposed FastRef model in Sec. \ref{sec: method} to three popular and recently developed prototype-oriented few-shot IAD methods, including PatchCore \cite{roth2022towards}, FastRecon \cite{fang2023fastrecon}, WinCLIP \cite{jeong2023winclip}, and AnomalyDINO \cite{damm2025anomalydino}. Next, we will discuss them one by one.

\subsection{PatchCore+: Application to PatchCore}
\label{sec: patchcore+}

PatchCore \cite{roth2022towards} is originally introduced for one-class IAD with a large number of normal images. Recently, however, it has also been applied to few-shot IAD, thanks to its flexible prototype-oriented design \cite{santos2023optimizing}. 
A notable drawback of PatchCore is that its prototypes remain fixed during inference, which causes it to overlook the characteristics of the query images. Therefore, we propose PatchCore+ by incorporating our proposed FastRef to enhance the representativeness of original prototypes in PatchCore.  Specifically,
we use Euclidean distance in line with PatchCore, and substitute it into Eq. \ref{eq: obj}, Eq. \ref{eq: ot}, and Eq. \ref{eq: det}, allowing us to rewrite the nested optimization problem and anomaly score map as:
\begin{align} \label{eq: patchcore+}
\begin{split}
    & \Wv^*, \Tv^* := \mathop{\arg\min}\limits_{\Wv, \Tv} \|\fv_{{\rm{t}}}^{{\rm{q}}}-\Wv \boldsymbol{\mathcal{M}}_{\rm{s}}\|^2+\lambda {\rm{OT}}(p,q) \\
    & \sv_{\rm{j}}:={\rm{min}}_{\boldsymbol{r} \in \boldsymbol{{\mathcal{M}}}^{\rm{w}*}_{\rm{s}}} \|\fv_{{\rm{t}, \rm{j}}}^{{\rm{q}}}- \boldsymbol{r}\|^2, \quad {\rm{j}}=1,...,{\rm{m}}
\end{split}
\end{align}
where $\boldsymbol{{\mathcal{M}}}^{\rm{w}*}_{\rm{s}}=\Wv^* \boldsymbol{{\mathcal{M}}}_{\rm{s}}$. And the few-shot IAD can then be carried out as described in Sec. \ref{sec: detection}.

\begin{figure*}[!t]
\centering
\centerline{\includegraphics[width=14cm]{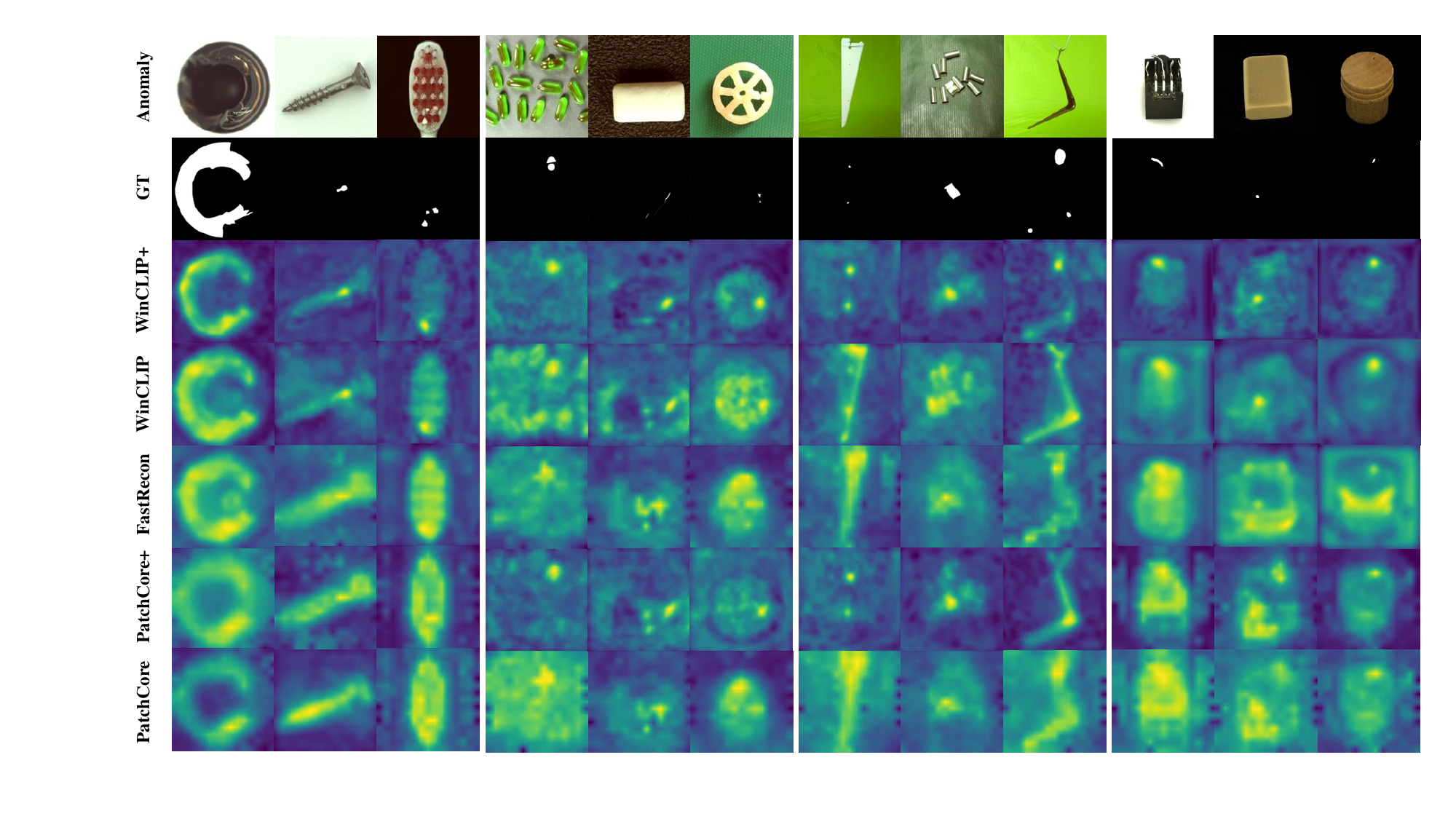}}
\caption{From left to right are results of pixel-level anomaly localization with MVTec, ViSA, MPDD, and RealIAD under 4-shots.}
\label{fig:visualization0}
\end{figure*}

\subsection{FastRecon+: Application to FastRecon}
\label{sec: fastrecon+}

FastRecon \cite{fang2023fastrecon} is a pioneer of fast prototype refinement as far as we know. However, they mainly focus on transferring characteristics from query features while pay less attention on suppressing anomalies in the transferred query features. Specifically, FastRecon only employs point-to-point regularization to enhance the normality of the refined prototypes $\Wv^* \boldsymbol{{\mathcal{M}}}_{\rm{s}}$, which is proven to be less effective in our experiments. Hence, we propose FastRecon+ by replacing the point-to-point regularization in FastRecon with the OT regularization, as shown in the second term of Eq. \ref{eq: obj}. We use the Euclidean distance as FastRecon does for ${\rm{dis}}(\cdot,\cdot)$ in Eq. \ref{eq: obj}. Additionally, the measurement function used in the cost matrix of Eq. \ref{eq: ot} is also the Euclidean distance. Moreover, the rule of our FastRecon+ for updating $\Wv$ is different from that of FastRecon. For details, please compare Eq. \ref{eq: updatew} of our model and Eq. 12 of FastRecon in \cite{fang2023fastrecon}.
We observe that the FastRecon+ is the same with the PatchCore+ in Sec. \ref{sec: patchcore+}.

\subsection{WinCLIP+: Application to WinCLIP}
\label{sec: winclip+}

Unlike PatchCore and FastRecon whose visual features are extracted from CNNs \cite{zagoruyko2016wide, tan2019efficientnet}
, WinCLIP \cite{jeong2023winclip} extracts visual features from a pre-trained CLIP-based large visual language model 
\cite{radford2021learning}. Although it demonstrates superior performance in few-shot IAD, it neglects the importance of efficiently transferring characteristics from query images. We propose WinCLIP+ by applying FastRef to WinCLIP, it uses cosine distance to rewrite Eq. \ref{eq: obj}, Eq. \ref{eq: ot}, and Eq. \ref{eq: det} as:
\begin{align*} \label{eq: winclip+}
\begin{split}
& \Wv^*, \Tv^* := \mathop{\arg\min}\limits_{\Wv, \Tv} \frac{1}{2} [1 - {\rm{cos}}( \fv_{\rm{t}}^{\rm{q}}, \Wv \boldsymbol{\mathcal{M}}_{\rm{s}})] +\lambda {\rm{OT}}(p,q) \\
    & \sv_{\rm{j}}:={\rm{min}}_{\boldsymbol{r} \in \boldsymbol{{\mathcal{M}}}^{\rm{w}*}_{\rm{s}}} \frac{1}{2} [1 - {\rm{cos}}( \fv_{\rm{t},\rm{j}}^{\rm{q}}, \boldsymbol{r})], \quad {\rm{j}}=1,...,{\rm{m}}
\end{split}
\end{align*}
where $\boldsymbol{{\mathcal{M}}}^{\rm{w}*}_{\rm{s}}=\Wv^* \boldsymbol{{\mathcal{M}}}_{\rm{s}}$. To conduct few-shot IAD for $\fv_{\rm{t}}^{\rm{q}}$, we combine the maximum value of $\sv$ above with the WinCLIP zero-shot anomaly score $s_0: \mathbb{R}^{\rm{c}} \rightarrow [0,1]$ \cite{jeong2023winclip}, which provide complementary information, one from query feature references and the other from CLIP knowledge by:
\begin{align}
\begin{split}
    s^* = \frac{1}{2}[s_0(\fv_{\rm{t}}^{\rm{q}})+{\rm{max}}_{{j} \in [1,{\rm{m}}]}\sv_j]
\end{split}
\end{align}

\subsection{AnomalyDINO+: Application to AnomalyDINO}
\label{sec: AnomalyDINO+}

AnomalyDINO \cite{damm2025anomalydino} employs more scalable foundation Transformer, DINOv2 \cite{oquab2023dinov2}, that extracts universal and discriminative features compared with WinCLIP. Using the normal prototypes (also called the memory bank in AnomalyDINO) and query features collected from AnomalyDINO, we propose AnomalyDINO+ by applying our FastRef to AnomalyDINO.
We use the same prototype refinement objective to obtain $\boldsymbol{\mathcal{M}}_{\rm{s}}^{w*}$
and anomaly score map for anomaly detection and localization
as those used in WinCLIP+.

\section{Experiments}
\label{sec: exp}

We conduct comprehensive experiments by applying our proposed FastRef to three prototype-oriented few-shot IAD methods including PatchCore, FastRecon, and WinCLIP discussed in Sec. \ref{sec: application} under 1-shots, 2-shots and 4-shots. We first evaluate both image- and pixel-level performance to demonstrate the effectiveness of FastRef. Then, ablation studies and hyper-parameters' impacts are implemented to validate the improvements of different components. 
Finally, we report the inference time to verify the efficiency.


\begin{table}[t]
    \centering
    \caption{
    Few-shot IAD performance on MVTec and MPDD. FR is abbreviation of FastRef. $\Delta$ is improvement. Results of image-level and pixel-level are reported in AUROC (\%). 
    The best and the second best results are bold with black and blue, respectively.
    }
    \captionsetup{}
\begin{center}
    \resizebox{0.6\textwidth}{!}{
    \begin{tabular}{c c c cc|cc}
        \toprule
        \multicolumn{1}{c}{\multirow{2}*{Setup}} &\multicolumn{1}{c}{\multirow{2}*{Method}} &\multicolumn{1}{c}{\multirow{2}*{FR}} & \multicolumn{2}{c}{MVTec} & \multicolumn{2}{c}{MPDD} \\ 
        \cmidrule{4-7}
        & & & Image & Pixel & Image & Pixel \\ 
        \midrule
        \multirow{18}{*}{1-shot} 
        & PaDiM (ICPR'21) & $\times$  &  74.1 & 87.5  & 47.3  & 87.7  \\
        & RegAD (ECCV'22) & $\times$ &  82.9 & 92.5  & 60.9  & 92.6  \\
        & GraphCore (ICLR'23) & $\times$ &  89.9 & \bb{\textbf{95.6}} & \textbf{84.7}  & 95.2  \\
        & FOCT (ACMMM'24) & $\times$ &  87.1 & 94.4 & \bb{\textbf{78.9}} & 96.2 \\
        & PromptAD (CVPR'24) & $\times$ &  92.9 & 95.1  & 73.1  & 95.1  \\
        \cmidrule{2-7}
        & \multirow{3}{*}{PatchCore (CVPR'22)}
        & $\times$ & 84.1 & 92.3 & 71.0 & 96.3 \\
        & & $\checkmark$ & 85.9 & 93.7 & 74.9 & \bb{\textbf{96.6}} \\
        & & $\Delta$ & +1.8 & +1.4 & +3.9 & +0.3 \\
        \cmidrule{2-7}
        & \multirow{3}{*}{FastRecon (ICCV'23)}
        & $\times$ & 85.7 & 93.2 & 74.1  & 96.3 \\
        & & $\checkmark$ & 85.9 & 93.7 & 74.9 & \bb{\textbf{96.6}} \\
        & & $\Delta$ & +0.2 & +0.5 & +0.8 & +0.3 \\
        \cmidrule{2-7}
        & \multirow{3}{*}{WinCLIP (CVPR'23)}
        & $\times$ & \bb{\textbf{93.5}} & 93.6 & 70.5 & 96.3 \\
        & & $\checkmark$ & \textbf{93.8} & \textbf{95.7} & 72.5 & \textbf{96.9} \\
        & & $\Delta$ & +0.3 & +2.1 & +2.0 & +0.6 \\
        \midrule
        \multirow{18}{*}{2-shot} 
        & PaDiM (ICPR'21) & $\times$ &  77.4 & 90.1 & 53.6  & 89.5 \\
        & RegAD (ECCV'22) & $\times$ &  85.7 & 94.6 & 63.4  & 93.2 \\
        & GraphCore (ICLR'23) & $\times$ &  91.9 & \textbf{96.9} & \textbf{85.4}  & 95.4 \\
        & FOCT (ACMMM'24) & $\times$ &  90.5 & 94.8 & \bb{\textbf{82.4}} & 96.5 \\
        & PromptAD (CVPR'24) & $\times$ &  93.4 & 95.4  & 80.1  & 95.8 \\
        \cmidrule{2-7}
        & \multirow{3}{*}{PatchCore (CVPR'22)}
        & $\times$ & 87.1 & 93.3  & 71.4 & 96.5 \\
        & & $\checkmark$ & 88.8 & 94.7 & 78.2 & \bb{\textbf{96.9}} \\
        & & $\Delta$ & +1.7 & +1.4 & +6.8 & +0.4 \\
        \cmidrule{2-7}
        & \multirow{3}{*}{FastRecon (ICCV'23)}
        & $\times$ & 88.3 & 94.5 & 76.4 & 96.7 \\
        & & $\checkmark$ & 88.8 & 94.7 & 78.2 & \bb{\textbf{96.9}} \\
        & & $\Delta$ & +0.5 & +0.2 & +1.8 & +0.2 \\
        \cmidrule{2-7}
        & \multirow{3}{*}{WinCLIP (CVPR'23)}
        & $\times$ & \bb{\textbf{93.7}} & 93.8  & 72.5 & 96.5 \\
        & & $\checkmark$ & \textbf{93.9} & \bb{\textbf{96.2}} & 76.0 & \textbf{97.3} \\
        & & $\Delta$ & +0.2 & +2.4 & +3.5 & +0.8 \\
        \midrule
        \multirow{18}{*}{4-shot} 
         & PaDiM (ICPR'21) & $\times$ &  79.8 & 92.1  & 56.1 & 90.6  \\
         & RegAD (ECCV'22) & $\times$ &  88.2 & 95.8  & 68.3 & 93.9  \\
         & GraphCore (ICLR'23) & $\times$ &  92.9 & \textbf{97.4}  & \textbf{85.7} & 95.7  \\
         & FOCT (ACMMM'24) & $\times$ & 93.2 & 96.2  & \bb{\textbf{83.2}} & 96.7 \\
         & PromptAD (CVPR'24) & $\times$ & \textbf{95.5} & 96.3  & 80.4 & 96.2  \\
        \cmidrule{2-7}
        & \multirow{3}{*}{PatchCore (CVPR'22)}
        & $\times$ & 90.0 & 95.1 & 76.2 & \bb{\textbf{97.2}} \\
        & & $\checkmark$ & 92.1 & 96.1  & 80.3 & \bb{\textbf{97.2}} \\
        & & $\Delta$ & +2.1 & +1.0  & +4.1 & +0.0 \\
        \cmidrule{2-7}
        & \multirow{3}{*}{FastRecon (ICCV'23)}
        & $\times$ & 91.3 & 96.1 & 79.7 & 96.9 \\
        & & $\checkmark$ & 92.1 & 96.1 & 80.3 & \bb{\textbf{97.2}} \\
        & & $\Delta$ & $+0.8$  & $+0.0$ & $+0.6$ & $+0.3$ \\
        \cmidrule{2-7}
        & \multirow{3}{*}{WinCLIP (CVPR'23)}
        & $\times$ & \bb{\textbf{95.3}} & 94.2  & 75.0 & 96.8 \\
        & & $\checkmark$ & \textbf{95.5} & \bb{\textbf{96.7}} & 82.0 & \textbf{97.6} \\
        & & $\Delta$ & +0.2 & +2.5 & +7.0 & +0.8 \\
        \bottomrule
    \end{tabular}}
    \end{center}
    \label{tab:MainResults-1}
\end{table}

\begin{table}[t]
    \centering
    \caption{Few-shot IAD performance on VisA and RealIAD. FR is abbreviation of FastRef. $\Delta$ is improvement. Results of image-level and pixel-level are reported in AUROC (\%). 
    The best and the second best results are bold with black and blue, respectively.}
    \captionsetup{}
\begin{center}
    \resizebox{0.68\textwidth}{!}{
    \begin{tabular}{c c c cc|cc}
        \toprule
        \multicolumn{1}{c}{\multirow{2}*{Setup}} &\multicolumn{1}{c}{\multirow{2}*{Method}} &\multicolumn{1}{c}{\multirow{2}*{FR}} & \multicolumn{2}{c}{ViSA} & \multicolumn{2}{c}{RealIAD} \\ 
        \cmidrule{4-7}
        & & & Image & Pixel & Image & Pixel \\ 
        \midrule
        \multirow{14}{*}{1-shot} 
        & PaDiM (ICPR'21) & $\times$  &  58.8 & 86.6  & 64.0 & 93.7  \\
        & PromptAD (CVPR'24) & $\times$ &  \textbf{86.5} & 96.2  & 67.8 & 91.6  \\
        \cmidrule{2-7}
        & \multirow{3}{*}{PatchCore (CVPR'22)}
        & $\times$ & 71.0 & 96.1 & 70.9 & 91.2 \\
        & & $\checkmark$ & 78.3 & \textbf{97.1}  &\textbf{75.9} &\textbf{96.3} \\
        & & $\Delta$ & +7.3 & +1.0 &+5.0 &+5.1 \\
        \cmidrule{2-7}
        & \multirow{3}{*}{FastRecon (ICCV'23)}
        & $\times$ & 76.2 & \bb{\textbf{96.7}}   & 71.2 & \bb{\textbf{95.7}} \\
        & & $\checkmark$ & 78.3 & \textbf{97.1}  &\textbf{75.9} &\textbf{96.3} \\
        & & $\Delta$ & +2.1 & +0.4  &+4.7 & +0.6 \\
        \cmidrule{2-7}
        & \multirow{3}{*}{WinCLIP (CVPR'23)}
        & $\times$ & 83.4 & 94.7 &73.8 &94.3 \\
        & & $\checkmark$ & \bb{\textbf{83.9}} & 95.8 &\bb{\textbf{74.4}} & 94.8 \\
        & & $\Delta$ & +0.5 & +1.1 &+0.6 &+0.5 \\
        \midrule
        \multirow{14}{*}{2-shot} 
        & PaDiM (ICPR'21) & $\times$ &  62.5 & 89.9  & 66.4 & 94.6 \\
        & PromptAD (CVPR'24) & $\times$ &  \bb{\textbf{86.7}} & 96.5  &74.5 & 93.5 \\
        \cmidrule{2-7}
        & \multirow{3}{*}{PatchCore (CVPR'22)}
        & $\times$ & 80.0 & 96.9  & 71.7 & 91.3 \\
        & & $\checkmark$ & \textbf{87.1} & \textbf{98.0}  &\textbf{76.9} &\textbf{96.5} \\
        & & $\Delta$ & +7.1 & +1.1  &+5.2 &+5.2 \\
        \cmidrule{2-7}
        & \multirow{3}{*}{FastRecon (ICCV'23)}
        & $\times$ & 86.1 & \bb{\textbf{97.6}}  & 72.5 & \bb{\textbf{95.9}} \\
        & & $\checkmark$ & \textbf{87.1} & \textbf{98.0}  &\textbf{76.9} &\textbf{96.5} \\
        & & $\Delta$ & +1.0 & +0.4  & +4.4 & +0.6 \\
        \cmidrule{2-7}
        & \multirow{3}{*}{WinCLIP (CVPR'23)}
        & $\times$ & 83.8 & 95.1  & 75.0 & 94.6 \\
        & & $\checkmark$ & 84.1 & 96.4 & \bb{\textbf{75.9}} & 95.2 \\
        & & $\Delta$ & +0.3 & +1.3 & +0.9 & +0.6 \\
        \midrule
        \multirow{14}{*}{4-shot} 
         & PaDiM (ICPR'21) & $\times$ &  65.5 & 92.3  & 69.4 & 95.3  \\
         & PromptAD (CVPR'24) & $\times$ & \bb{\textbf{88.8}} & 96.8  & \textbf{78.5} & 94.9  \\
        \cmidrule{2-7}
        & \multirow{3}{*}{PatchCore (CVPR'22)}
        & $\times$ & 84.2 & 97.5 &75.9& 93.0 \\
        & & $\checkmark$ & \textbf{90.4} & \textbf{98.2} &\bb{\textbf{77.4}} & \textbf{96.7} \\
        & & $\Delta$ & +6.2 & +0.7 & +1.5 & +3.7 \\
        \cmidrule{2-7}
        & \multirow{3}{*}{FastRecon (ICCV'23)}
        & $\times$ & 88.2 & \bb{\textbf{98.0}}  & 73.2 & \bb{\textbf{96.0}} \\
        & & $\checkmark$ & \textbf{90.4} & \textbf{98.2} &\bb{\textbf{77.4}} & \textbf{96.7} \\
        & & $\Delta$ & +2.2 & +0.2 & +4.2 & +0.7 \\
        \cmidrule{2-7}
        & \multirow{3}{*}{WinCLIP (CVPR'23)}
        & $\times$ & 84.1 & 95.4  & 76.4 & 94.8 \\
        & & $\checkmark$ & 85.0 & 96.6 &77.3 & 95.3 \\
        & & $\Delta$ & +0.9 & +1.2 & +0.9 & +0.5 \\
        \bottomrule
    \end{tabular}}
    \end{center}
    \label{tab:MainResults-2}
\end{table}

\subsection{Experiment Setup}

\textbf{(1) Datasets.} 
We conduct experiments on MVTec \cite{bergmann2019mvtec}, VisA \cite{zou2022spot}, MPDD \cite{jezek2021deep}, and RealIAD \cite{wang2024real} datasets.
The MVTec dataset consists of 3,629 training images and 1,725 test images across 15 categories, covering 5 types of textures and 10 types of objects, with each category exhibiting an average of five distinct defect types. Image resolutions ranging from 700×700 to 1,024×1,024. The VisA dataset contains 9,621 normal images and 1,200 anomaly images featuring 78 types of anomalies. It is divided into 12 subsets, each representing a distinct object, with an average of 6.5 defect types per subset. Image resolutions are around 1,500×1,000. The MPDD dataset includes 888 normal training images and 458 test images, including 176 normal and 282 abnormal images, spanning 6 classes of metal products with a resolution of 1,024×1,024. 
The RealIAD dataset provides over 150K images covering 30 different classes. It also provides multi-view images with pixel-level annotations. There are totally 99721 normal images and 51329 abnormal images with a resolution of up to 2,000$\sim$5,000.

\noindent \textbf{(2) Competing Methods.}
We compare our models described in Sec. \ref{sec: application} with some most recently proposed few-shot IAD methods or those applicable in low-data regimes, including PaDiM \cite{defard2021padim}, RegAD \cite{huang2022registration}, PatchCore \cite{roth2022towards}, FastRecon \cite{fang2023fastrecon}, 
WinCLIP \cite{jeong2023winclip},
GraphCore \cite{xie2023pushing},
FOCT \cite{tian2024foct},
and PromptAD \cite{li2024promptad}.
We report results on RealIAD by implementing experiments with the official codes of PaDiM, PromptAD, PatchCore, FastRecon, and WinCLIP. 

\noindent \textbf{(3) Evaluation Protocols.} 
We evaluate the performance of anomaly detection and localization using image/pixel-level AUROC \cite{jeong2023winclip,li2024promptad}. Additionally, we assess real-time efficiency by measuring the running time per image at test time.

\noindent \textbf{(4) Implementation Details.}
For PatchCore+ and FastRecon+, we use the pre-trained WRN-50 \cite{zagoruyko2016wide} to extract features from the intermediate layers, following \cite{roth2022towards, fang2023fastrecon}. Images are all resized to 256×256. 
Balanced coefficient $\lambda=0.3$ and Coreset sampling ratio $\alpha=0.05$. 
As for WinCLIP+, we resize all images to 240$\times$240 and use the pre-trained CLIP model with ViT-B/16+ to extract features following \cite{jeong2023winclip}. 
We use the smallest model of DINOv2 \cite{oquab2023dinov2} in implementing our developed AnomalyDINO+, and evaluate input resolution with 448 pixels following \cite{damm2025anomalydino}.
$\lambda=0.1$ and $\alpha=0.5,0.3,0.2,0.1$ for MVTec-AD, VisA, MPDD, and RealIAD datasets. 
At the test time, the batch size equals to 1 to isolate the performance gains from our method.
All experiments are conducted on a NVIDIA GTX 3090 GPU.

\begin{table}[t]
    \centering
    \captionsetup{width=0.9\textwidth}
    \caption{Ablation studies of WinCLIP+ with image and pixel AUROCs (\%) under 2-shots. The best results are in bold.}
    \label{tab: ablation}
    \vspace{0.5em}
    \resizebox{0.68\textwidth}{!}{%
    \begin{tabular}{c c cc|cc|cc}
        \toprule
        \multicolumn{1}{c}{\multirow{2}*{$\Wv^*$}} &
        \multicolumn{1}{c}{\multirow{2}*{$\Tv^*$}} &
        \multicolumn{2}{c}{MVTec} &
        \multicolumn{2}{c}{VisA} &
        \multicolumn{2}{c}{MPDD} \\
        \cmidrule{3-8}
        & & Image & Pixel & Image & Pixel & Image & Pixel \\
        \midrule
        \textbf{$\times$} & \textbf{$\times$} & 93.7 & 93.8 & 83.8 & 95.1 & 72.5 & 96.5 \\
        \textbf{$\checkmark$} & \textbf{$\times$} & 93.8 & 94.7 & 84.0 & 96.2 & 74.9 & 96.8 \\
        \textbf{$\checkmark$} & \textbf{$\checkmark$} & \textbf{93.9} & \textbf{96.2} & \textbf{84.1} & \textbf{96.4} & \textbf{76.0} & \textbf{97.3} \\
        \bottomrule
    \end{tabular}}
\end{table}

\begin{figure}[t]
    \centering
    \includegraphics[width=0.68\textwidth]{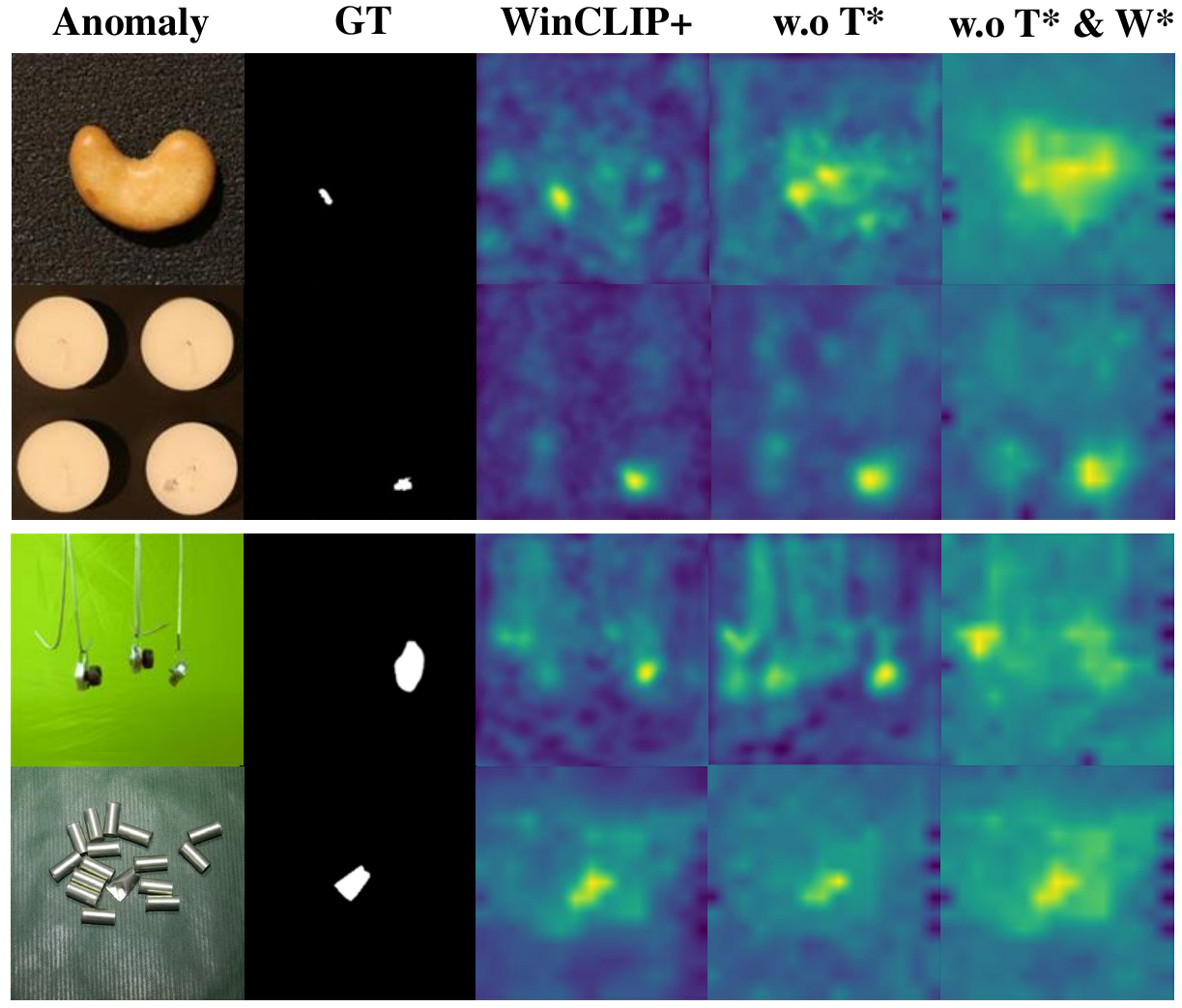}
    \captionsetup{width=0.9\textwidth}
    \caption{From top to bottom are results on ablation studies of 2-shots anomaly localization on MVTec and VisA datasets.}
    \label{fig: ablation}
\end{figure}

\subsection{Comparisons with SOTA Methods}

\textbf{(1) Image-level Comparison Results.} 
We present image-level anomaly detection results in Table \ref{tab:MainResults-1} and Table \ref{tab:MainResults-2}. PaDiM \cite{defard2021padim} and PatchCore \cite{santos2023optimizing} are adapted from traditional full-shot methods to few-shot settings.
Comparing the results of RegAD \cite{huang2022registration}, PatchCore \cite{santos2023optimizing}, and FastRecon \cite{fang2023fastrecon}, the following observations are evident: i) Prototype-oriented methods outperform the meta-learning models, demonstrating the superior flexibility of their feature representations; ii) FastRecon significantly outperforms PatchCore, highlighting the importance of incorporating characteristics from query images. 
When comparing PatchCore and FasRecon to WinCLIP \cite{jeong2023winclip}, WinCLIP achieves a substantial performance gain except on the MPDD dataset. We attribute this to two factors: i) Generally, the representations from pre-trained CLIP models are more powerful than those from CNNs; ii) Unlike MVTec, VisA and RealIAD, MPDD is a metal dataset with rotation variations that may not be well-represented during CLIP pre-training. This discrepancy motivates us to demonstrate the effectiveness of our model using both CNN-based and CLIP-based pre-trained models. By comparing the results of PatchCore/WinCLIP with PatchCore+/WinCLIP+, we observe that PatchCore+/WinCLIP+ consistently delivers superior IAD performance, indicating that our prototype refinement is effective. For example, WinCLIP+ achieves a 7\% improvement in AUROC on the MPDD dataset under 4-shots. Furthermore, the improvement delivered by our model surpasses that of the point-to-point regularization approach used in FastRecon \cite{fang2023fastrecon}, underscoring the importance of refining prototypes in a more systematic way. Additionally, as we have discussed in Sec. \ref{sec: application} that PatchCore+ and FastRecon+ are the same, thus the results of them are also the same and we omit to analyze FastRecon+ here.

\noindent \textbf{(2) Pixel-level Comparison Results.} 
Pixel-level anomaly localization results are presented in Table \ref{tab:MainResults-1} and Table \ref{tab:MainResults-2}.
When comparing PatchCore+ and WinCLIP+ with other methods, the trend in pixel-level AUROC is consistent with that of the image-level AUROC. It suggests that our model not only effectively detects anomalous images but also accurately localizes the anomalous regions. We attribute these performance gains to our well-refined prototypes.

\begin{figure}[!t]
\centering
\centerline{\includegraphics[width=9cm]{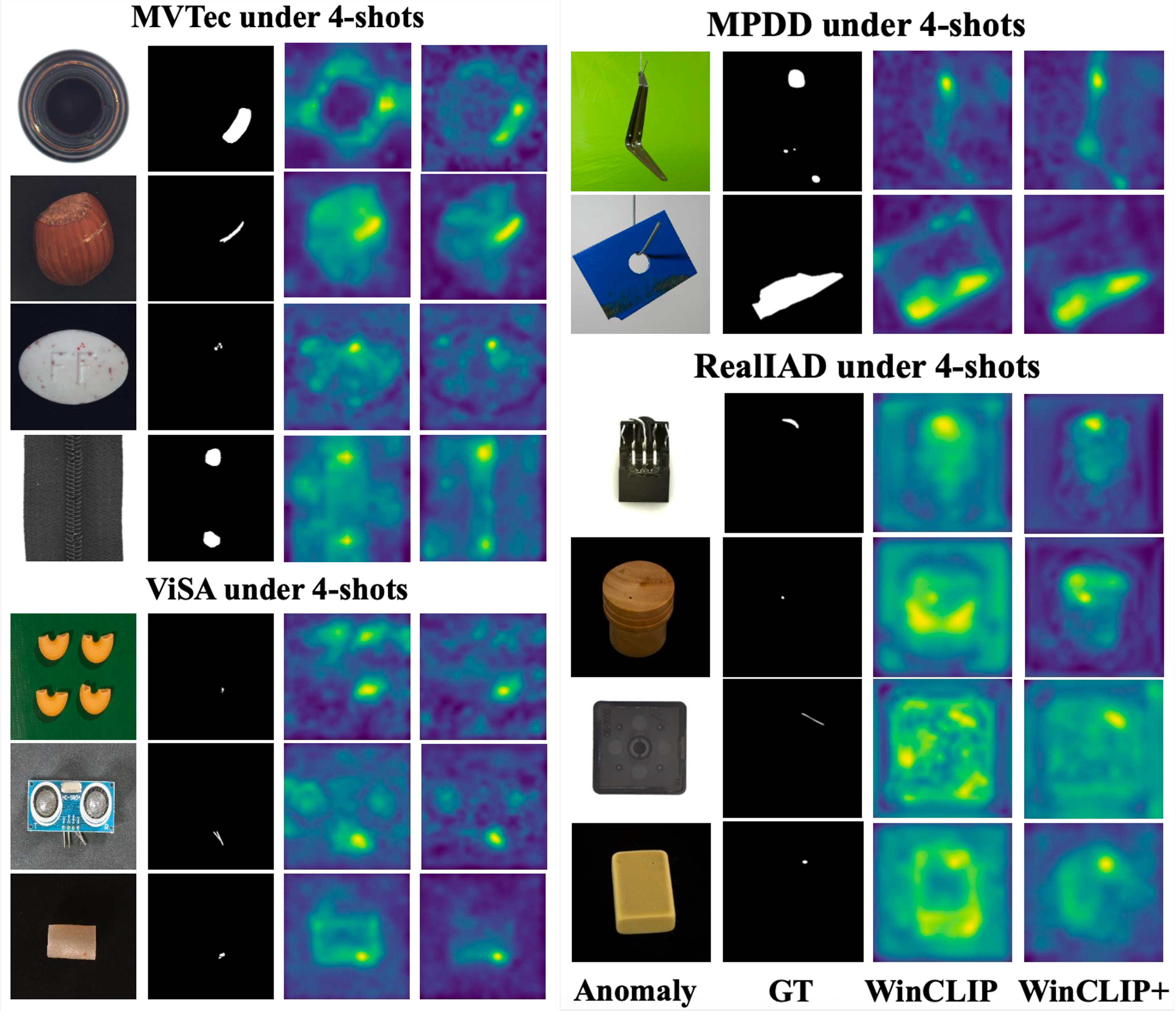}}
\vspace{-2mm}
\caption{More visualization results of pixel-level anomaly localization on MVTec, ViSA, MPDD, and RealIAD under 4-shots.}
\label{fig:visualization}
\end{figure}

\noindent \textbf{(3) Qualitative results.}
Visualizations for all the datasets versus different methods are shown in Fig. \ref{fig:visualization0}. WinCLIP+ achieves more precise anomaly localization, particularly for subtle anomalies. 
This further validate the effectiveness of our proposed FastRef model and its flexibility for applying to prototype-oriented few-shot IAD methods. More results of WinCLIP versus WinCLIP+ are reported in Fig. \ref{fig:visualization}.



\subsection{Ablative Analysis}

We use WinCLIP+ to assess the impact of each module in our proposed FastRef, including anomaly suppression via the transport probability $\Tv^*$ and characteristic transfer through the transform matrix $\Wv^*$ under 2-shots on three datasets
. The results are reported in Table \ref{tab: ablation}, with visualizations on the datasets of MVTec and VisA shown in Fig. \ref{fig: ablation}, where ${\rm{w.o}} \ \Tv^*$ means WinCLIP+ without $\Tv^*$, ${\rm{w.o}} \ \Tv^*  \&  \Wv^*$ denotes WinCLIP+ without $\Tv^*$ and $\Wv^*$.

\noindent \textbf{(1) Impact of Anomaly Suppression by $\Tv^*$.}
Anomalies in query images negatively affect robust anomaly detection due to the limited diversity and representativeness of normal prototypes. Therefore, leveraging query images through effective anomaly suppression is crucial. As shown in Table \ref{tab: ablation}, 
anomaly suppression yields averaged improvements of over $0.4\%$ in image-level AUROC while over $0.7\%$ in pixel-level AUROC on all the three datasets.

\noindent \textbf{(2) Impact of Characteristic Transfer $\Wv^*$.} 
According to Table \ref{tab: ablation}, $\Wv^*$ consistently improves detection and localization performance. However, in same cases we observe that the gains from using $\Tv^*$ are more pronounced than those from $\Wv^*$. For example, image-level detection on the MPDD dataset, the gain from using $\Wv^*$ is $2.4\%$, whereas further using $\Tv^*$ results in a $1.1\%$ improvement. This suggests that relying solely on $\Wv^*$ may introduce anomalies, thereby limiting overall performance improvements.

\begin{table}[!t]
  \caption{FS-IAD performance versus backbones in image-level and \textcolor{gray}{pixel-level} AUROC (\%) $\uparrow$ on MPDD dataset under 2-shots.}
  \vspace{-4mm}
\begin{center}
  \resizebox{0.65\textwidth}{!}{
  \begin{tabular}{c|c|c}
    \toprule
    \multicolumn{1}{c|}{\multirow{2}*{Backbones}}
    & \multicolumn{1}{c|}{PatchCore}  & \multicolumn{1}{c}{PatchCore+} \\
& {(CVPR'22)} & {(Ours)}   \\
    \midrule
EfficientNet\_b5     &62.5 {\textcolor{gray}{/ 87.6}}  &{69.3 (+6.8)} {\textcolor{gray}{/ {92.9 (+5.3)}}} \\
ViT\_base     &62.7 {\textcolor{gray}{/ 88.4}}  &{71.6 (+8.9)} {\textcolor{gray}{/ {92.2 (+3.8)}}} \\
WRN50     &71.4 {\textcolor{gray}{/ 96.5}}  &{78.2 (+6.8)} {\textcolor{gray}{/ {96.9 (+0.4)}}} \\
\bottomrule
  \end{tabular}}
  \end{center}
  \label{tablebackbone}
   \vspace{-1mm}
\end{table}

\begin{table}[!t]
\vspace{-3mm}
  \caption{FS-IAD performance of TTT+ (Gaussian) and OT (non-Gaussian) combined with PatchCore for anomaly suppression in image/\textcolor{gray}{pixel}-level AUROC (\%) $\uparrow$ on MPDD dataset.}
  \vspace{-4mm}
\begin{center}
  \resizebox{0.58\textwidth}{!}{
  \begin{tabular}{c|c|c}
    \toprule
    \multicolumn{1}{c|}{\multirow{2}*{Shot}}
    & \multicolumn{1}{c|}{TTT+}  & \multicolumn{1}{c}{OT} \\
& {(NeurIPS'21)} & {(Ours)}   \\
    \midrule
1     &74.2 {\textcolor{gray}{/ 96.3}}  &{74.9}(+0.7) {\textcolor{gray}{/ {96.6}(+0.3)}} \\
2     &76.8 {\textcolor{gray}{/ 96.5}}   &{78.2} (+1.4){\textcolor{gray}{/ {96.9}(+0.4)}} \\
4     &79.8 {\textcolor{gray}{/ 96.6}}   &{80.3}(+0.5) {\textcolor{gray}{/ {97.2}(+0.6)}} \\
\bottomrule
  \end{tabular}}
  \end{center}
  \label{table2ttt}
\end{table}

\begin{figure*}[!t]
\centering
\centerline{\includegraphics[width=15cm]{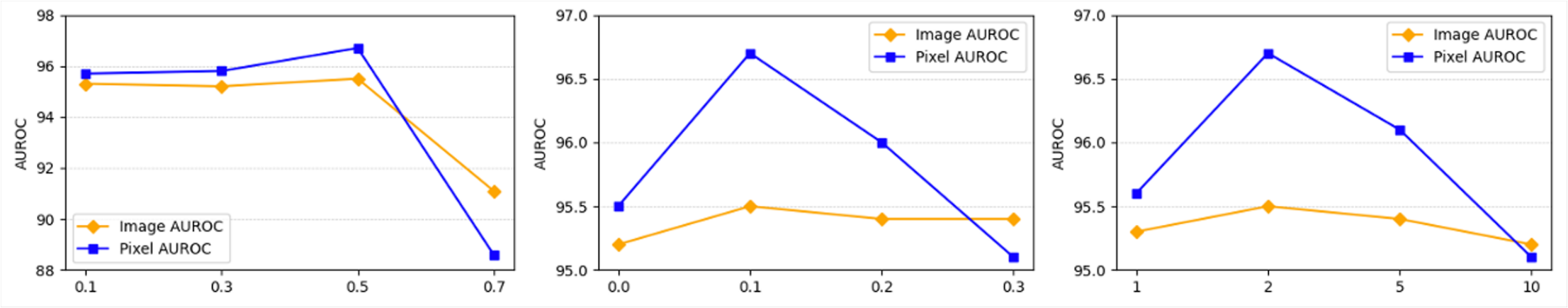}}
\caption{From left to right are analysis on hyper-parameters of $\alpha$, $\lambda$, and iteration number with MVTec dataset under 4-shots.
}
\label{fig: hyper}
\end{figure*}

\begin{figure*}[t]
    \centering
    \begin{minipage}[t]{0.5\textwidth}
        \centering
        \includegraphics[width=\textwidth]{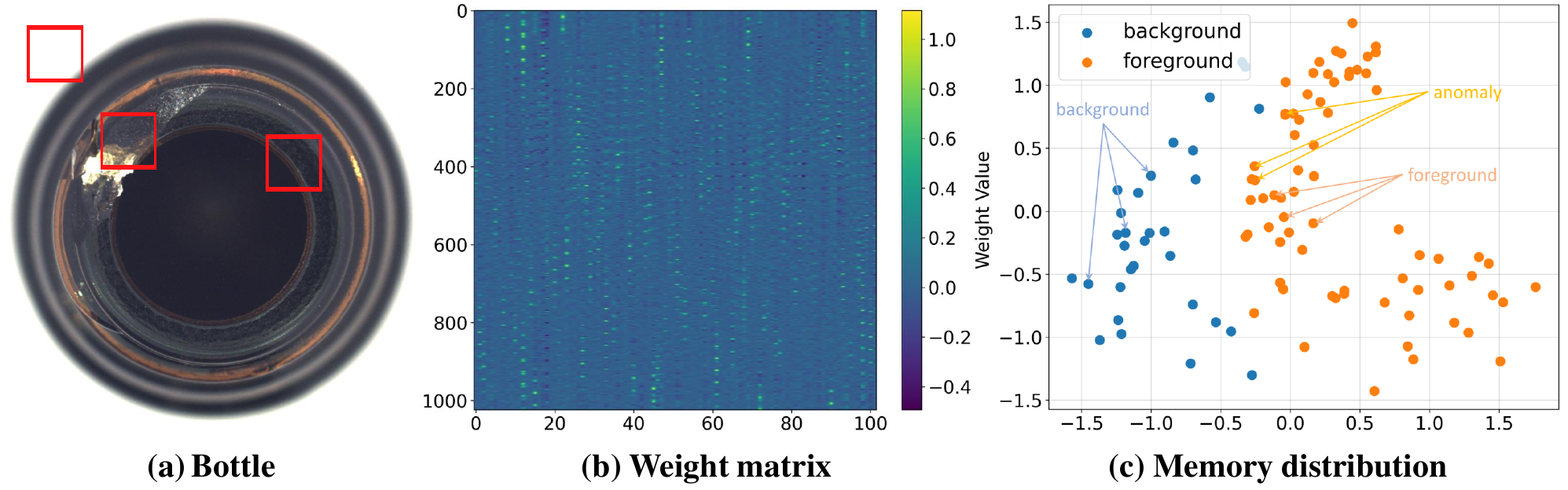} 
        \vspace{-1em} 
        \captionsetup{width=7cm}
        \caption{The transform matrix in (b) corresponding to the bottle in (a). (c) shows the 2-D PCA of Top-3 selected items in the normal prototypes according to (b) for foreground, background, and anomaly regions of query image in (a).}
        \label{fig: visW}
    \end{minipage}
    \hfill
    \begin{minipage}[t]{0.48\textwidth}
        \vspace{-6em} 
        \captionsetup{width=7cm}
        \captionof{table}{Inference time (s) per image on MVTec.}
        \centering
        \resizebox{0.85\textwidth}{!}{
        \begin{tabular}{c|c}\toprule
            Methods & Inference time \\ \midrule
            PromptAD (CVPR'24) & 0.257 \\
            PatchCore (CVPR'22) & 0.098 \\
            PatchCore+ (ours) & 0.100 \\ 
            FastRecon (ICCV'23) & 0.099 \\
            FastRecon+ (ours) & 0.100 \\
            WinCLIP (CVPR'23) & 0.180  \\
            WinCLIP+ (ours) & 0.184 \\
            \bottomrule
        \end{tabular}}
        \label{tab: time}
    \end{minipage}
\end{figure*}

\textbf{Remark:} 
Both $\Tv^*$ and $\Wv^*$ play important roles in boosting the performance of few-shot IAD, which also echoes analysis in Fig. \ref{fig:rule}. Next, we try to explain a counterintuitive phenomenon that using normal prototypes can actually reconstruct abnormal patterns in few-shot IAD, which could never be encountered in PatchCore when the normal images are abundant \cite{roth2022towards}. The phenomenon can be obviously observed when comparing results between WinCLIP+ and w.o $\boldsymbol{T}^* \& \boldsymbol{W}^*$ (WinCLIP) according to Table \ref{tab: ablation} and Fig. \ref{fig: ablation}. We attribute this to the fact that the diversity of normal features is poor in few-shot IAD, thus, it makes the model more prone to shortcut compared with model trained in scenarios with sufficient normal images.
To better understand this, we visualize the transform matrix $\boldsymbol{W}^*$ using WinCLIP+ on an anomalous bottle image, as shown in Fig. \ref{fig: visW}.
In this example, $\Wv^* \in \mathbb{R}^{1024 \times 102}$, where 1024 is the number of query features given the input bottle image, and 102 is the number of prototypes. Each row of $\Wv^*$ is sparse, the value of which represents the strength of current query feature selecting a specific item in the normal prototypes $\boldsymbol{{\mathcal{M}}}_{\rm{s}}$, it also echos with our claim of  composition refinement on the transform matrix in Sec. \ref{sec: formulation}. Furthermore, we visualize $\Wv^*$ corresponding to three different regions in the image of the bottle marked with a red box that includes the foreground region, the background region, and the anomaly region, the results are reported in Fig. \ref{fig: visW} (c). Taking the foreground region $\Wv^*_{{\rm{f}}} \in \mathbb{R}^{102}$ for example, we select the highest Top-3 values in $\Wv^*_{{\rm{f}}}$ and use their indexes to select the corresponding prototypes in $\boldsymbol{{\mathcal{M}}}_{\rm{s}}$. Then we visualize these selected prototypes using PCA on a 2-D space. For better view, we draw all prototypes in the $\boldsymbol{{\mathcal{M}}}_{\rm{s}}$ on the 2-D space and use different color to indicate prototypes that belonging to foreground or background. As we can see that, for query features on foreground region, it also tends to select foreground prototypes. Similar analysis could be conducted on background region and anomaly region. Interestingly, we find that the query feature in the anomaly region also prefers to select foreground prototypes, thus reconstructing the anomalies, which is consistent with our analysis above and verifies the necessary use of $\Tv^*$ to suppress anomalies in the low-data regime.

\subsection{Hyper-parameters Analysis}

We report results of hyperparameters' impacts using WinCLIP+ under 4-shots setting on the MVTec dataset in Fig. \ref{fig: hyper}
, including the Coreset sampling ratio $\alpha$, the balanced coefficient $\lambda$, and {the iteration number ${\rm{L}}$.}

\noindent \textbf{(1) Impact of CoreSet sampling ratio $\alpha$.}
As shown in Fig. \ref{fig: hyper} (left), 
We observe that the Coreset sampling ratio $\alpha$ is crucial in determining IAD performance, as it controls the initial representativeness of the prototypes.

\noindent \textbf{(2) Impact of the balanced coefficient $\lambda$.}
According to the results of Fig. \ref{fig: hyper} (middle),
We observe a similar trend as in $\alpha$ that the performance first improves and then declines as $\lambda$ changes. This phenomenon suggests that characteristic transfer should dominate the success of prototype refinement, aligning with our design.

\noindent {\textbf{(3) Impact of the iteration number $\rm{L}$.} 
The refined prototypes are obtained in an iterative manner, thus, the iteration number ${\rm{L}}$ is crucial for robust few-shot IAD. Results reported in Fig. \ref{fig: hyper} (right) indicate that the WinCLIP+, enhanced by our FastRef, achieves strong IAD performance when ${\rm{L}}=2$, demonstrating the efficiency of our model.}

\subsection{More Analysis}

\noindent \textbf{(1) Analysis on the real-time efficiency}.
The running time at test time is reported in Table \ref{tab: time}. Although the most recently proposed PromptAD is competitive in anomaly detection as shown in Table \ref{tab:MainResults-1} and Table \ref{tab:MainResults-2}, it costs a little more running time per image compared with other methods due to its complex prompt learning process. 
Additionally, our FastRef-enhanced methods including PatchCore+, FastRecon+ and WinCLIP+ achieves competitve inference time and satisfies real-time requirement in practice, which benefits from efficiency of our method discussed in Sec. \ref{sec: solution}.

\noindent \textbf{(2) Analysis on the robustness against various backbones.}
We report FS-IAD results versus different backbones in Table \ref{tablebackbone}. Two observations are drawn: i) backbones show different basic performance, and we attribute this to their architectures and pre-training; ii) our method achieves consistent improvements across all the selected backbones, indicating its robustness and superiority.

\noindent \textbf{(3) Analysis on the non-Gaussian assumption}.
OT-based anomaly suppression technique introduced in Sec. \ref{sec: solution} assumes that the refined prototypes and original (normal) prototypes are non-Gaussian-like, which could be suitable for the cases where the feature distributions are non-Gaussian or unknown. To verify the soundness and advantages quantitatively, we report FS-IAD performance using different anomaly suppression technique with Gaussian distribution assumption, as shown in Table \ref{table2ttt}. Specifically, we use a statistic alignment method, called TTT+ \cite{liu2021ttt++}, developed for test time training (TTT) \cite{sun2020test}. It assumes feature distribution is Gaussian-like, and uses Gaussian statistics of source and target domains to decrease gaps for distribution alignment. In FS-IAD application, the source and target domains are separately represented by the refined prototypes $\boldsymbol{\mathcal{M}}^{\rm{w}}_{\rm{s}}$ and their original normal counterparts $\boldsymbol{\mathcal{M}}_{\rm{s}}$, the means of which are denoted by $\boldsymbol{\mu}_{\boldsymbol{\mathcal{M}}^{\rm{w}}_{\rm{s}}}$ and $\boldsymbol{\mu}_{\boldsymbol{\mathcal{M}}_{\rm{s}}}$. Therefore, in TTT+ case, we directly replace our OT-based anomaly suppression term in Eq. \ref{eq: obj} with the statistic alignment term as:
\begin{align} \label{eq:ttt}
\begin{split}
    {\rm{Align}}(p,q)=\|\boldsymbol{\mu}_{\boldsymbol{\mathcal{M}}^{\rm{w}}_{\rm{s}}}-\boldsymbol{\mu}_{\boldsymbol{\mathcal{M}}_{\rm{s}}}\|_2^2
\end{split}
\end{align}
At this moment, transform matrix $\boldsymbol{W}$ is the only unknown parameter required to be solved. Following the similar derivation as in Sec. \ref{sec: solution}, a one-step solution for $\boldsymbol{W}^*$ with closed can be obtained. few-shot IAD performance can then be derived by substituting $\boldsymbol{W}^*$ into the anomaly detection method introduced in Sec. \ref{sec: detection}. The formulation and the derivation of the one-step $\boldsymbol{W}^*$ can be found in the Appendix.
According to the above analysis, we empirically demonstrate that our OT-based anomaly suppression with non-Gaussian assumption should be more robust and practical versus its Gaussian-like counterpart.
Additionally, our method can also be regarded as a special case of TTT in the few-shot IAD regime, which has rarely been studied before.

\section{Conclusion}
This paper addresses the problem of few-shot IAD by introducing a fast prototype refinement framework (FastRef), which can be easily integrated with any prototype-oriented FS-IAD methods. We first formulate FastRef as a nested optimization problem for characteristic transfer and anomaly suppression, and then propose an efficient iterative algorithm to solve the parameters in a closed form. The convergence of FastRef is also guaranteed.
The experiments on four widely used datasets have validated the effectiveness and efficiency of our method.
We hope that our work can provide some novel insights for a broader range of prototype-oriented few-shot anomaly detection applications beyond IAD for social safety.

\appendix

\section{Derivation of Eq. \ref{eq: updatew}}
When we employ Euclidian or cosine distance in Eq. \ref{eq: obj}, we first take Euclidean distance and present the derivation of Eq. \ref{eq: updatew} in updating transform matrix $\boldsymbol{W}$.
Then we explain that, whether it is Euclidean distance or cosine distance, the way of updating the transform matrix should be the same.
Substituting the Euclidean distance into the distance function $\rm{dis(.,.)}$ in Eq. \ref{eq: obj}, for the $j$-th query features of the $t$-th query image $\boldsymbol{f}_t^q(j,:)$, the objective can be re-written as:
\begin{align*} \footnotesize
\begin{split}
    \mathcal{L}_{\rm{i}}&=\|\boldsymbol{f}_{\rm{t}}^{\rm{q}}({\rm{i}},:)-\boldsymbol{W}_{{\rm{i}},:}\boldsymbol{\mathcal{M}}_{\rm{s}}\|_2^2 + \lambda \sum_{{\rm{j}}=1}^{\rm{n}} \boldsymbol{T}_{{\rm{i}},{\rm{j}}} \|\boldsymbol{W}_{{\rm{i}},:}\boldsymbol{\mathcal{M}}_{\rm{s}}-\boldsymbol{\mathcal{M}}_{\rm{s}}({\rm{j}},:)\|_2^2 \\
    &\propto(1+\lambda \sum_{{\rm{j}}=1}^{\rm{n}} \boldsymbol{T}_{{\rm{i}},{\rm{j}}})\boldsymbol{W}_{{\rm{i}},:}\boldsymbol{\mathcal{M}}_s\boldsymbol{\mathcal{M}}_s^T\boldsymbol{W}_{{\rm{i}},:}^T \quad \quad \quad {\rm{quadratic}} \ {\rm{term}} \\
    & \quad -2\boldsymbol{W}_{{\rm{i}},:}\boldsymbol{\mathcal{M}}_s[\boldsymbol{f}_{\rm{t}}^{\rm{q}}({\rm{i}},:)+\lambda \sum_{{\rm{j}}=1}^{\rm{n}} \boldsymbol{T}_{{\rm{i}},{\rm{j}}}\boldsymbol{\mathcal{M}}_{\rm{s}}({\rm{j}},:)]^T \quad {\rm{linear}} \ {\rm{term}}
\end{split}
\end{align*}
where $i$ denotes the $i$-th item in the normal prototypes $\boldsymbol{\mathcal{M}}_s$. $\boldsymbol{W}$ and $\boldsymbol{T}$ separately are transform matrix and transport probability discussed in Sec. \ref{sec: formulation}. As we can see that this is a quadratic form in terms of $\boldsymbol{W}_{j,:}$, according to \cite{petersen2008matrix}:
\begin{align*} \footnotesize
\begin{split}
    \frac{\partial \mathcal{L}_{\rm{i}}}{\partial \boldsymbol{W}_{{\rm{i}},:}}&=[2(1+\lambda \sum_{{\rm{j}}=1}^{\rm{n}} \boldsymbol{T}_{{\rm{i}},{\rm{j}}})\boldsymbol{\mathcal{M}}_{\rm{s}}\boldsymbol{\mathcal{M}}_s^T\boldsymbol{W}^T_{{\rm{i}},:}]^T \\
    &- \{2\boldsymbol{\mathcal{M}}_{\rm{s}}[\boldsymbol{f}_{\rm{t}}^{\rm{q}}({\rm{i}},:)+\lambda \sum_{{\rm{j}}=1}^{\rm{n}} \boldsymbol{T}_{{\rm{i}},{\rm{j}}}\boldsymbol{\mathcal{M}}_{\rm{s}}({\rm{j}},:)]^T\}^T
\end{split}
\end{align*}
Then the optimal solution for $\boldsymbol{W}_{j,:}$ can be computed by setting $\frac{\partial \mathcal{L}}{\partial \boldsymbol{W}_{j,:}}=0$, and we have:
\begin{align*} \footnotesize
\begin{split}
    \boldsymbol{W}_{{\rm{i}},:}=\frac{[\boldsymbol{f}_{\rm{t}}^{\rm{q}}({\rm{i}},:)\boldsymbol{\mathcal{M}}_{\rm{s}}^T+\lambda \sum_{{\rm{j}}=1}^{\rm{n}} \boldsymbol{T}_{{\rm{i}},{\rm{j}}}\boldsymbol{\mathcal{M}}_{\rm{s}}({\rm{j}},:)\boldsymbol{\mathcal{M}}_{\rm{s}}^T](\boldsymbol{\mathcal{M}}_{\rm{s}}\boldsymbol{\mathcal{M}}_{\rm{s}}^T)^{-1}}{1+\lambda \sum_{i=1}^{n} \boldsymbol{T}_{i,j}}
\end{split}
\end{align*}
It is obvious that the rows of $\boldsymbol{W}$ can be updated in parallel and expressed in matrix form as follows:
\begin{align*} \footnotesize
\begin{split}
    \boldsymbol{W}=\frac{(\boldsymbol{f}_t^q\boldsymbol{\mathcal{M}}_s^T+\lambda \boldsymbol{T}\boldsymbol{\mathcal{M}}_s\boldsymbol{\mathcal{M}}_s^T)(\boldsymbol{\mathcal{M}}_s\boldsymbol{\mathcal{M}}_s^T)^{-1}}{1+\lambda \boldsymbol{T} \cdot \textbf{1}}
\end{split}
\end{align*}

Next, we gonna illustrate that the optimal solution above is also suitable when we choose cosine distance in Eq. \ref{eq: obj}. Taking the first term in Eq. \ref{eq: obj} as an example while the second one is the same, given $\boldsymbol{f}_{\rm{t}}^{\rm{q}}({\rm{i}},:)$, $\boldsymbol{\mathcal{M}}_{\rm{s}}$, and $\boldsymbol{W}_{\rm{i},:}$, considering the fact that $\boldsymbol{f}_{\rm{t}}^{\rm{q}}({\rm{i}},:)$ and $\boldsymbol{\mathcal{M}}_{\rm{s}}$ are known and fixed during updating $\boldsymbol{W}_{{\rm{i}},:}$, hence, it is obvious that both Euclidean distance and cosine distance (defined in Sec. \ref{sec: winclip+}) achieve minimum when $\boldsymbol{f}_{\rm{t}}^{\rm{q}}({\rm{i}},:)=\boldsymbol{W}_{{\rm{i}},:}\boldsymbol{\mathcal{M}}_{\rm{s}}$. Therefore, the optimal $\boldsymbol{W}^*$ should be the same for employing either Euclidean distance or cosine distance.  
$\hfill\blacksquare$

\section{Derivation of the One-step Solution for $\boldsymbol{W}^*$}
We replace the second term of OT-based anomaly suppression with Eq. \ref{eq:ttt}, and re-write the objective in Eq. \ref{eq: obj} as:
\begin{align*} \footnotesize
\begin{split}
    \mathcal{L}_{\rm{i}}& =\|\boldsymbol{f}_{\rm{t}}^{\rm{q}}({\rm{i}},:)-\boldsymbol{\mathcal{M}}^{\rm{w}}_{{\rm{s}},{\rm{i}}}\|_2^2 + \lambda \|\boldsymbol{\mu}_{{\boldsymbol{\mathcal{M}}}_{\rm{s}}} - \frac{\sum_{{\rm{i}}=1}^{{\rm{m}}}\boldsymbol{\mathcal{M}}^{\rm{w}}_{{\rm{s}},{\rm{i}}}}{{\rm{m}}}\|_2^2 
\end{split}
\end{align*}
where $\boldsymbol{\mathcal{M}}^{\rm{w}}_{{\rm{s}},{\rm{i}}}=\boldsymbol{W}_{{\rm{i}},:}\boldsymbol{\mathcal{M}}_{\rm{s}}$, $\boldsymbol{W}_{{\rm{i}},:}$ is the ${\rm{i}}$-th row of $\boldsymbol{W}$. As we can see that this is a quadratic form in terms of $\boldsymbol{W}_{{\rm{i}},:}$, 
for the two terms, we separately express them as:
\begin{align*} \footnotesize
\begin{split}
    \mathcal{L}_{\rm{i}}^{1}& =\|\boldsymbol{f}_{\rm{t}}^{\rm{q}}({\rm{i}},:)-\boldsymbol{\mathcal{M}}^{\rm{w}}_{{\rm{s}},{\rm{i}}}\|_2^2 \\
    & = (\boldsymbol{f}_{\rm{t}}^{\rm{q}}({\rm{i}},:)-\boldsymbol{W}_{{\rm{i}},:}\boldsymbol{\mathcal{M}}_{{\rm{s}}})(\boldsymbol{f}_{\rm{t}}^{\rm{q}}({\rm{i}},:)-\boldsymbol{W}_{{\rm{i}},:}\boldsymbol{\mathcal{M}}_{{\rm{s}}})^T\\
    & \propto -2 \boldsymbol{W}_{{\rm{i}},:}\boldsymbol{\mathcal{M}}_{\rm{s}}[\boldsymbol{f}_{\rm{t}}^{\rm{q}}({\rm{i}},:)]^T + \boldsymbol{W}_{{\rm{i}},:}\boldsymbol{\mathcal{M}}_{\rm{s}}\boldsymbol{\mathcal{M}}^{T}_{\rm{s}}\boldsymbol{W}^{T}_{{\rm{i}},:} \\
    \mathcal{L}_{\rm{i}}^{2}& =\lambda \|\boldsymbol{\mu}_{{\boldsymbol{\mathcal{M}}}_{\rm{s}}} - \frac{\sum_{{\rm{i}}=1}^{{\rm{m}}}\boldsymbol{\mathcal{M}}^{\rm{w}}_{{\rm{s}},{\rm{i}}}}{{\rm{m}}}\|_2^2 \\
    & = \lambda(\boldsymbol{\mu}_{{\boldsymbol{\mathcal{M}}}_{\rm{s}}}-\frac{\sum_{{\rm{i}}=1}^{{\rm{m}}}\boldsymbol{W}_{{\rm{i}},:}\boldsymbol{\mathcal{M}}_{{\rm{s}}}}{{\rm{m}}})(\boldsymbol{\mu}_{{\boldsymbol{\mathcal{M}}}_{\rm{s}}}-\frac{\sum_{{\rm{i}}=1}^{{\rm{m}}}\boldsymbol{W}_{{\rm{i}},:}\boldsymbol{\mathcal{M}}_{{\rm{s}}}}{{\rm{m}}})^T\\
    & \propto \lambda [-2 \frac{\boldsymbol{W}_{{\rm{i}},:}\boldsymbol{\mathcal{M}}_{\rm{s}}\boldsymbol{\mu}^T_{\boldsymbol{\mathcal{M}}_{\rm{s}}}}{m} + \frac{\sum_{{\rm{r}}\neq i}\boldsymbol{W}_{{\rm{r}},:}\boldsymbol{\mathcal{M}}_{\rm{s}}\boldsymbol{\mathcal{M}}^T_{\rm{s}} \boldsymbol{W}^T_{{\rm{i}},:}}{{\rm{m}}^2} \\
    & \quad +\frac{\boldsymbol{W}_{{\rm{i}},:}\boldsymbol{\mathcal{M}}_{\rm{s}}\boldsymbol{\mathcal{M}}^T_{\rm{s}} \boldsymbol{W}^T_{{\rm{i}},:}}{{\rm{m}}^2}] \\
\end{split}
\end{align*}

The two terms above are quadratic forms in terms of $\boldsymbol{W}_{{\rm{i}},:}$, according to \cite{petersen2008matrix}, we separately have:
\begin{align*} \footnotesize
\begin{split}
    \frac{\partial \mathcal{L}^{1}_{\rm{i}}}{\partial \boldsymbol{W}_{{\rm{i}},:}}&=-\{2 \boldsymbol{\mathcal{M}}_{\rm{s}}[\boldsymbol{f}_{\rm{t}}^{\rm{q}}({\rm{i}},:)]^T\}^T+2[\boldsymbol{\mathcal{M}}_{\rm{s}}\boldsymbol{\mathcal{M}}^T_{\rm{s}}\boldsymbol{W}^T_{{\rm{i}},:}]^T\\
    \frac{\partial \mathcal{L}_{\rm{i}}^{2}}{\partial \boldsymbol{W}_{{\rm{i}},:}}&=-2[\frac{\boldsymbol{\mathcal{M}}_{\rm{s}}\boldsymbol{\mu}^T_{\boldsymbol{\mathcal{M}}_{\rm{s}}}}{\rm{m}}]^T+\frac{\sum_{{\rm{r}}\neq {\rm{i}}}\boldsymbol{W}_{{\rm{r}},:}\boldsymbol{\mathcal{M}}_{\rm{s}}\boldsymbol{\mathcal{M}}^T_{\rm{s}}}{{\rm{m}}^2}+\frac{2[\boldsymbol{\mathcal{M}}_{\rm{s}}\boldsymbol{\mathcal{M}}^T_{\rm{s}}\boldsymbol{W}^T_{{\rm{i}},:}]^T}{{\rm{m}}^2}
\end{split}
\end{align*}
Then the optimal solution for $\boldsymbol{W}_{{\rm{i}},:}$ can be computed by setting $\frac{\partial \mathcal{L}_{\rm{i}}^{1}+\mathcal{L}_{\rm{i}}^{2}}{\partial \boldsymbol{W}_{{\rm{i}},:}}=0$, and we have:
\begin{align*} \footnotesize
\begin{split}
    \boldsymbol{W}_{{\rm{i}},:}=\frac{[2{\rm{m}}^2\boldsymbol{f}_{\rm{t}}^{\rm{q}}({\rm{i}},:)+2{\rm{m}}\boldsymbol{\mu}_{\boldsymbol{\mathcal{M}}_{\rm{s}}}-\sum_{{\rm{r}}\neq{\rm{i}}}\boldsymbol{W}_{\rm{r}}\boldsymbol{\mathcal{M}}_{\rm{s}}]\boldsymbol{\mathcal{M}}^T_{\rm{s}}(\boldsymbol{\mathcal{M}}_{\rm{s}}\boldsymbol{\mathcal{M}}^T_{\rm{s}})^{-1}}{2+2{\rm{m}}^2}
\end{split}
\end{align*}
As we can see that the rows of $\boldsymbol{W}$ can be updated in parallel and rewritten by:
\begin{align*} \tiny
\begin{split}
    \boldsymbol{W}=\frac{[2{\rm{m}}^2\boldsymbol{f}_{\rm{t}}^{\rm{q}}+2{\rm{m}}\boldsymbol{\mu}^{\rm{Rep}}_{\boldsymbol{\mathcal{M}}_{\rm{s}}}-(\sum_{{\rm{r}}=1}^{\rm{m}}\boldsymbol{W}_{\rm{r}}\boldsymbol{\mathcal{M}}_{\rm{s}})^{\rm{Rep}}+\boldsymbol{W}\boldsymbol{\mathcal{M}}_{\rm{s}}]\boldsymbol{\mathcal{M}}^T_{\rm{s}}(\boldsymbol{\mathcal{M}}_{\rm{s}}\boldsymbol{\mathcal{M}}^T_{\rm{s}})^{-1}}{2+2{\rm{m}}^2}
\end{split}
\end{align*}
where the superscript ${\rm{Rep}}$ indicates that the vector dimensions are replicated to align with $\boldsymbol{W}$'s dimensions.
$\hfill\blacksquare$

\section{Proof of Convergence}
We can rewrite the objective in Eq. \ref{eq: obj} by substituting ${\rm{dis}}(\cdot,\cdot)$ with Euclidean distance and ${\rm{OT}}(\cdot,\cdot)$ with Eq. \ref{eq: ot} as follows:
\begin{align*} \footnotesize
\begin{split}
    \mathcal{L}=&F(\boldsymbol{W})+\lambda G(\boldsymbol{W},\boldsymbol{T})\\
    =&\sum_{{\rm{i}}=1}^{\rm{m}}\|\boldsymbol{f}_{\rm{t}}^{\rm{q}}({\rm{i}},:)-\boldsymbol{W}_{{\rm{i}},:}\boldsymbol{\mathcal{M}}_{\rm{s}}\|_2^2 \\
    & + \lambda \sum_{{\rm{i}}=1}^{\rm{m}}\sum_{{\rm{j}}=1}^{\rm{n}} \boldsymbol{T}_{{\rm{i}},{\rm{j}}} \|\boldsymbol{W}_{{\rm{i}},:}\boldsymbol{\mathcal{M}}_{\rm{s}}-\boldsymbol{\mathcal{M}}_{\rm{s}}({\rm{j}},:)\|_2^2
\end{split}
\end{align*}
where $\boldsymbol{T}$ satisfies the regularization of $\boldsymbol{T}\in \mathcal{T}:=\{\boldsymbol{T}\in \mathbb{R}^{{\rm{m}}\times{\rm{n}}}|\boldsymbol{T}_{{\rm{i}},{\rm{j}}}\geq0,\sum_{{\rm{i}}=1}^{m} \boldsymbol{T}_{{\rm{i}},{\rm{j}}}=\frac{1}{\rm{n}},\sum_{{\rm{j}}=1}^{n} \boldsymbol{T}_{{\rm{i}},{\rm{j}}}=\frac{1}{\rm{m}}\}$
. According to \cite{bubeck2015convex}, we can see that $F(\boldsymbol{W})$ is $\mu$-strongly convex and L-smooth with respect to (w.r.t) $\boldsymbol{W}$ while $G(\boldsymbol{W},\boldsymbol{T})$ is linear w.r.t $\boldsymbol{T}$, therefore, the sub-problems of updating $\boldsymbol{W}$ and $\boldsymbol{T}$ are separately strong convex optimization and linear programming, which can be formally expressed as:
\begin{align*} \footnotesize
\begin{split}
    \boldsymbol{W}_{l+1}&=\underset{\boldsymbol{W}}{\arg\min}\mathcal{L}(\boldsymbol{W},\boldsymbol{T}_l)\\
    &=\underset{\boldsymbol{W}}{\arg\min}[F(\boldsymbol{W})+\lambda G(\boldsymbol{W},\boldsymbol{T}_l)] \\
    \boldsymbol{T}_{l+1}&=\underset{\boldsymbol{T}\in \mathcal{T}}{\arg\min}\mathcal{L}(\boldsymbol{W}_{l+1},\boldsymbol{T})=\underset{\boldsymbol{T}\in \mathcal{T}}{\arg\min} G(\boldsymbol{W},\boldsymbol{T}_l)
\end{split}
\end{align*}

Since $F(\boldsymbol{W})+\lambda G(\boldsymbol{W},\boldsymbol{T}_l)$ is $\mu$-strongly convex and $\boldsymbol{T}_{l+1}$ minimizes $G(\boldsymbol{W}_{l+1},\boldsymbol{T})$, we have:
\begin{align*} \footnotesize
\begin{split}
    & \mathcal{L}(\boldsymbol{W}_{l+1},\boldsymbol{T}_l) \leq \mathcal{L}(\boldsymbol{W}_l,\boldsymbol{T}_l)-\frac{\mu}{2} \|\boldsymbol{W}_{l+1}-\boldsymbol{W}_l\|_{\rm{F}}^2 \\
    & \mathcal{L}(\boldsymbol{W}_{l+1},\boldsymbol{T}_{l+1}) \leq \mathcal{L}(\boldsymbol{W}_{l+1},\boldsymbol{T}_l)
\end{split}
\end{align*}
Combining the above two inequalities, we have:
\begin{align*} \footnotesize
\begin{split}
    \mathcal{L}(\boldsymbol{W}_{l+1},\boldsymbol{T}_{l+1}) \leq \mathcal{L}(\boldsymbol{W}_l,\boldsymbol{T}_l)-\frac{\mu}{2} \|\boldsymbol{W}_{l+1}-\boldsymbol{W}_l\|_{\rm{F}}^2
\end{split}
\end{align*}
Therefore, the objective of optimizing our proposed FastRef should be convergent.
$\hfill\blacksquare$

\bibliographystyle{plainnat}
\bibliography{main}

\end{document}